\documentclass[11pt]{article}

\PassOptionsToPackage{table}{xcolor}

\usepackage[preprint]{acl}

\usepackage{times}
\usepackage{latexsym}
\usepackage[T1]{fontenc}
\usepackage[utf8]{inputenc}
\usepackage{microtype}
\usepackage{graphicx}
\usepackage{booktabs}
\usepackage{longtable}
\usepackage{amsfonts}
\usepackage{amsmath}
\usepackage{nicefrac}
\usepackage{enumitem}
\usepackage{array}
\usepackage{multirow}
\usepackage{listings}
\usepackage{newunicodechar}
\usepackage[most]{tcolorbox}
\usepackage{tikz}
\usetikzlibrary{arrows.meta,positioning,calc}
\lstdefinestyle{xalphacode}{
  language=Python,
  basicstyle=\scriptsize\ttfamily,
  keywordstyle=\color{blue!60!black},
  stringstyle=\color{teal!70!black},
  commentstyle=\color{gray!70},
  showstringspaces=false,
  frame=single,
  breaklines=true,
  columns=fullflexible,
  captionpos=b,
  xleftmargin=1em,
  framexleftmargin=1em,
  backgroundcolor=\color{gray!5},
  literate=
    {–}{{-}}1
    {-}{{-}}1
    {—}{{---}}1
    {ε}{{$\varepsilon$}}1
}
\newtcolorbox{xpromptbox}[1]{
  colback=gray!5,
  colframe=gray!80,
  title style={fontupper=\bfseries\small},
  title={#1},
  breakable
}
\newtcblisting{xpromptlisting}[1]{
  colback=gray!5,
  colframe=gray!80,
  title style={fontupper=\bfseries\small},
  title={#1},
  breakable,
  listing only,
  listing options={
    language={},
    basicstyle=\scriptsize\ttfamily\color{black},
    keywordstyle=\color{black},
    stringstyle=\color{black},
    commentstyle=\color{black},
    breaklines=true,
    columns=fullflexible,
    keepspaces=true,
    showstringspaces=false
  },
}

\definecolor{xalphaHeader}{HTML}{EEF2F7}
\definecolor{xalphaBest}{HTML}{BFD7FF}
\definecolor{xalphaSecond}{HTML}{D8E8FF}
\definecolor{xalphaThird}{HTML}{ECF5FF}
\definecolor{xalphaOursRow}{HTML}{FFF7E6}

\newcommand{\bestcell}[1]{\cellcolor{xalphaBest}\textbf{#1}}
\newcommand{\secondcell}[1]{\cellcolor{xalphaSecond}\underline{#1}}
\newcommand{\thirdcell}[1]{\cellcolor{xalphaThird}#1}

\newcommand{\method}{\textsc{X-Alpha}}

\title{\method: An AI Quant Researcher with Memory-Driven Self-Evolution}

\author{%
    \textbf{
    Fengyuan Liu$^{1,2}$ \hspace{0.5em}
    Yuchen Fu$^{1}$ \hspace{0.5em}
    Yuqi Wang$^{1}$ \hspace{0.5em}
    Fengyuan Liu$^{3}$ \hspace{0.5em}
    Mukai Li$^{1}$ \hspace{0.5em}
    Qi Liu$^{1,2}$
    }\\[3mm]
    $^{1}$The University of Hong Kong \quad
    $^{2}$Grace Investment Machine \quad
    $^{3}$Purdue University \\
    \small{\texttt{oxfengyuan@gmail.com}}
}

\begin{document}

\maketitle

\begin{abstract}
Human quantitative researchers mine alpha through an iterative process: they derive hypotheses from financial knowledge, implement and test them as factors, and use the resulting evidence to guide subsequent research.
Yet automated alpha mining faces three key bottlenecks: noisy financial knowledge is not directly actionable, exploration can remain confined to similar mechanisms, and feedback may remain tied to individual mining runs.
To address these bottlenecks, we introduce \method{}, an AI Quant Researcher that couples persistent research memory with continuous hypothesis-to-code discovery.
Report-to-Memory Abstraction (RMA) uses a multi-layer filtering-and-abstraction process to convert financial reports into actionable, data-compatible research paths.
Memory-driven routing uses accumulated experience to redirect exploration and define a targeted search space for each cycle.
Within each cycle, hypotheses are implemented as executable factors and refined through empirical validation and code-based evolution.
Empirical outcomes are consolidated into persistent research memory to guide subsequent exploration, enabling memory-driven self-evolution across cycles.
We evaluate \method{} on five equity indices spanning mainland China, the United States, and Hong Kong, with comprehensive experiments centered on CSI300.
Experiments show strong alpha discovery performance against baselines, with improvements in factor quality and signal diversity and consistent generation-level gains across multiple markets and prediction horizons.
\end{abstract}

\vspace{-0.2cm}
\section{Introduction}
\vspace{-0.2cm}
Alpha mining is the process of discovering financial factors that predict future returns from historical market observations.
This task is challenging because the search space is vast, financial markets are noisy and non-stationary~\citep{engle1982autoregressive}, and effective discovery requires more than formula search.
Human quantitative researchers address this challenge by drawing on financial knowledge to formulate hypotheses and implementing them as factors for empirical testing.
They use the resulting evidence to refine their research directions and accumulate experience for subsequent investigations.
This iterative process motivates our central question: how can an AI Quant Researcher sustain hypothesis-to-code alpha discovery through accumulated research experience?

Existing approaches include manually designed factors~\citep{fama1992cross,fama1993common,kakushadze2016101}, neural predictive models~\citep{duan2022factorvae,xu2021hist,xu2021rest}, and automated factor search using evolutionary algorithms or reinforcement learning~\citep{allen1999genetic,zhang2020autoalpha,cui2021alphaevolve,yu2023generating}.
More recently, LLM-based systems have introduced knowledge-guided factor construction and multi-agent research workflows~\citep{wang2023alphagpt,li2024can,tang2025alphaagent,shi2025alphajungle,li2025rdagent,wang2026factorminer,liu2026cognitive}.
Despite these advances, a continuous AI Quant Researcher still faces three key bottlenecks.
External financial knowledge often lacks actionable abstraction, exploration can remain confined to similar mechanisms, and feedback may not become reusable experience across research cycles.
Overcoming them requires reliable research abstraction, adaptive search-space control, and persistent knowledge accumulation.

\textbf{Knowledge abstraction bottleneck.}
External financial knowledge is usually noisy and not directly structured for factor research.
Reports mix actionable market mechanisms with details that are irrelevant, unavailable under the target data contract, or too coarse to guide factor implementation.
For example, analyst forecast revisions require data beyond daily OHLCV, and a generic momentum narrative may not specify an implementable research path.
Using reports as raw context can therefore misdirect exploration and obscure useful research clues.
We address this bottleneck with \textbf{Report-to-Memory Abstraction (RMA)}, a multi-layer filtering-and-abstraction process.
RMA distills noisy reports into actionable, data-compatible research paths without prescribing factor formulas.

\textbf{Search-space myopia.}
LLM-based factor generation can produce superficially different candidates that rely on similar financial mechanisms.
For example, varying a momentum factor's lookback window can produce different formulas within the same trend-continuation mechanism.
Candidate-level feedback alone provides limited guidance on when to move beyond an already explored research direction.
We address this bottleneck with \textbf{memory-driven routing}, a feedback-conditioned global search abstraction with two complementary functions.
First, it uses accumulated experience to redirect exploration and formulate the next cycle theme.
Second, it defines a targeted search space around this theme, retrieving relevant knowledge to guide hypothesis construction.

\textbf{Non-cumulative research.}
Existing systems can use feedback from successful and failed candidates to improve an ongoing mining process.
However, such feedback may remain tied to the current run rather than become reusable experience for long-term research.
We address this bottleneck with a closed research loop that consolidates empirical outcomes into persistent research memory.
Subsequent cycles use this memory to revise research directions and adapt search behavior.
Here, self-evolution refers to feedback-driven updates to memory and research behavior, not changes to the underlying LLM parameters.

Inspired by CogAlpha~\citep{liu2026cognitive}, we translate financial hypotheses into executable factors and refine them through code-based evolution.
We strengthen factor validation to assess whether implementations match their hypotheses and remain robust across evaluation periods.
Correlation-based screening further limits redundancy in the factor pool.
Report-grounded memory and research routing guide this discovery loop, and its empirical outcomes are consolidated into experience for subsequent cycles.
Together, these components form \textbf{\method{}}, a memory-driven AI Quant Researcher for continuous hypothesis-to-code discovery.
Experiments support RMA's predictive gains and memory-driven routing's improvement in cross-cycle signal diversity and quality.

The main contributions of this paper are:
\begin{itemize}[leftmargin=*, itemsep=2pt, topsep=2pt, parsep=0pt]
\item We develop \textbf{Report-to-Memory Abstraction (RMA)} to convert financial reports into actionable research paths and \textbf{memory-driven routing} to redirect exploration and define cycle-specific search spaces using accumulated experience.

\item We propose \textbf{\method{}}, an AI Quant Researcher that couples research memory with hypothesis-to-code discovery, enabling continuous research and memory-driven self-evolution across cycles.

\item We evaluate \method{} on five equity indices---CSI300, CSI500, CSI1000, SP500, and HSI90---across three markets, with comprehensive CSI300 experiments assessing factor quality and the effectiveness of our methods.
\end{itemize}

\vspace{-0.2cm}
\section{Related Work}
\label{sec:related}
\vspace{-0.2cm}

\textbf{LLM-based alpha mining.}
LLM-based alpha mining uses language models to interpret financial ideas and guide factor construction.
AlphaGPT and FAMA connect natural-language knowledge to interpretable alpha generation~\citep{wang2023alphagpt,li2024can}, while AlphaAgent and AlphaJungle improve symbolic factor search through regularized generation, tree search, and empirical feedback~\citep{tang2025alphaagent,shi2025alphajungle}.
Beyond factor construction, several systems also retain research experience to guide subsequent exploration.
RD-Agent(Q) organizes factor and model development within an iterative research loop~\citep{li2025rdagent}.
Its feedback primarily improves the research workflow, rather than explicitly defining cycle-specific search spaces.
FactorMiner retains successful patterns and forbidden regions to guide global formula search across mining sessions~\citep{wang2026factorminer}.
Its accumulated experience centers on formula search rather than the abstraction of external reports into actionable research paths.
CogAlpha evolves executable Python factors and reuses candidate feedback across generations and sub-cycles~\citep{liu2026cognitive}.
It does not explicitly connect report-grounded knowledge accumulation with cycle-level research routing.
In contrast, \method{} organizes report-grounded knowledge into actionable research paths.
Its memory-driven routing uses accumulated experience to redirect exploration across cycles and define theme-specific search spaces for hypothesis generation within each cycle.

\textbf{Autonomous and self-evolving research agents.}
Beyond alpha mining, LLM agents are increasingly studied as autonomous scientific researchers.
The AI Scientist exemplifies end-to-end research automation, taking scientific ideas through empirical investigation to a completed report~\citep{lu2024aiscientist}.
Multi-agent systems also explore how research is organized, from collaborative hypothesis development in AI Co-Scientist to sustained research cycles in EvoScientist~\citep{gottweis2026towards,lyu2026evoscientist}.
Other work examines how research agents learn from experience and acquired knowledge to adapt their behavior~\citep{qu2026coral,yan2026openskill,lee2026meta}.
These systems establish a broader paradigm of autonomous, iterative research.
However, alpha discovery requires financial hypotheses to be expressed as data-compatible factor code and evaluated for predictive value.
\method{} connects hypothesis-to-code discovery with financial consistency checks and empirical screening, providing a validated basis for subsequent research.

\vspace{-0.2cm}
\section{Approach}
\label{sec:approach}
\vspace{-0.2cm}

\begin{figure*}[t]
\centering
\includegraphics[width=\textwidth]{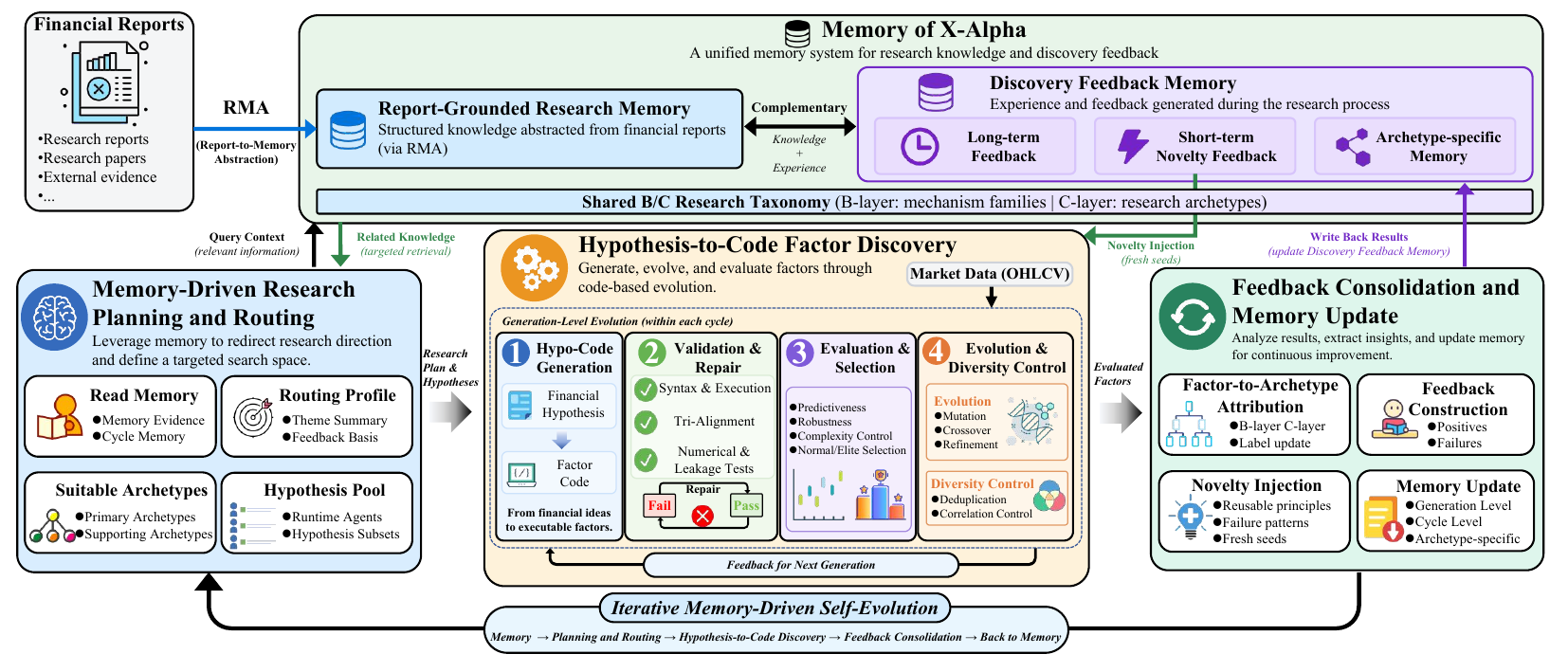}
\vspace{-0.8cm}
\caption{\textbf{Overall framework of \method{}.}
RMA-derived research knowledge and discovery feedback guide routing to redirect exploration and define cycle-specific search spaces.
Within each cycle, hypotheses are implemented and evolved as executable factors through empirical evaluation.
Research outcomes are consolidated into memory to guide subsequent cycles, closing the memory-driven self-evolution loop.}
\label{fig:xalpha_framework}
\vspace{-0.6cm}
\end{figure*}

\method{} is an AI Quant Researcher that couples persistent research memory with a hypothesis-to-code discovery loop to support continuous alpha research.
As shown in Figure~\ref{fig:xalpha_framework}, \textbf{RMA} supplies report-grounded research memory, complemented by discovery feedback from prior generations and cycles.
At the start of each cycle, memory-driven routing uses accumulated experience to redirect exploration and define a targeted search space.
Relevant knowledge is retrieved within this space to construct the hypothesis pool.
Hypotheses are implemented as executable factor code and refined through empirical validation and code-based evolution.
Factor-level outcomes are then consolidated into discovery feedback memory under the shared research taxonomy.
The updated memory guides subsequent cycles, enabling memory-driven self-evolution through changes in research directions and search behavior.

\begin{figure}[t]
\centering
\includegraphics[width=\linewidth]{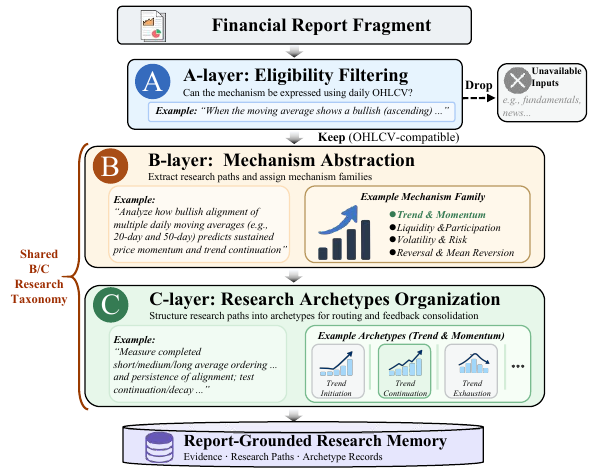}
\vspace{-0.6cm}
\caption{\textbf{RMA.}
The A-layer filters report evidence for OHLCV compatibility.
The B-layer extracts research paths by mechanism family.
The C-layer organizes and consolidates them under \emph{Research Archetypes}, forming reusable research memory.}
\label{fig:rma_abc_taxonomy}
\vspace{-0.5cm}
\end{figure}

\vspace{-0.2cm}
\subsection{Report-to-Memory Abstraction}
\label{sec:rma_layer}
\vspace{-0.2cm}

When financial reports are used directly as research context, useful clues remain mixed with information that depends on unavailable data.
Even data-compatible passages may be too broad to guide hypothesis construction.
To address this problem, inspired by structured-memory construction in StructMem~\citep{xu2026structmem}, we design Report-to-Memory Abstraction (RMA) as a multi-layer process that distills report content into actionable research memory under the daily OHLCV contract.

As shown in Figure~\ref{fig:rma_abc_taxonomy} above, RMA comprises three processing layers.
The \textbf{A-layer (eligibility filtering)} determines whether a report fragment contains a self-contained mechanism that can be studied using daily OHLCV data.
For example, the illustrated bullish moving-average alignment can be examined using daily prices and is therefore eligible for further processing.
The \textbf{B-layer (mechanism abstraction)} extracts concise, evidence-grounded research paths from eligible fragments.
Each path is assigned to a broad financial mechanism family.
In the figure, the moving-average discussion becomes a research question about whether bullish alignment supports sustained price momentum, assigned to \emph{Trend \& Momentum}.
The \textbf{C-layer (research archetype organization)} assigns each accepted path to a compatible archetype within its B-layer family.
Within each archetype, semantic de-duplication retains representative paths, followed by curation that consolidates closely related paths into more abstract research descriptions.
In the example, the moving-average path is assigned to the \emph{Trend Continuation} archetype.
Its curated memory record focuses on moving-average ordering and alignment persistence, asking whether the trend continues or decays.

The resulting memory provides data-compatible research paths that guide hypothesis construction within the selected research direction.
The predefined B-layer mechanism families and C-layer research archetypes form a shared research taxonomy (Appendix~\ref{app:archetype_taxonomy}).
Factor-to-archetype attribution maps code-validated candidate mechanisms to the same taxonomy for feedback construction (Section~\ref{sec:feedback_memory_update}).
Implementation details and recorded LLM-call examples are provided in Appendix~\ref{app:rma_details}.

\vspace{-0.2cm}
\subsection{Memory-Driven Routing}
\label{sec:memory_routing}
\vspace{-0.2cm}

As shown in Figure~\ref{fig:xalpha_framework}, memory-driven routing controls research at the cycle level: it redirects exploration using accumulated experience and defines a targeted search space before hypotheses are constructed.
For \textbf{research direction selection}, the planner reviews recent cycle feedback to assess which mechanisms have been explored and what the results reveal.
It uses this experience to identify an unresolved research question and formulate the next cycle theme.
For \textbf{search-space definition}, the planner maps this theme to primary and supporting mechanism families.
Relevant \emph{Research Archetypes} and their retrieved report-grounded paths provide the contexts for active research agents to construct the hypothesis pool.
A subset of these hypotheses is selected for seed-factor generation.

For example, earlier themes examined liquidity recovery after volume shocks.
Feedback highlighted an unresolved question about the relationship between post-spike volume-decay curvature and delayed price drift.
The planner redirected the next cycle theme toward this relationship.
The new search space was anchored in \emph{Volume Level and Structure}, with lagged-response archetypes connecting volume decay to later price adjustment.
Prior feedback guided the inclusion of volatility and liquidity conditions in the supporting research paths.
Implementation details and recorded examples appear in Appendix~\ref{app:memory_routing}.

\vspace{-0.2cm}
\subsection{Hypothesis-to-Code Factor Discovery}
\label{sec:hypothesis_to_code}
\vspace{-0.2cm}

As shown in Figure~\ref{fig:xalpha_framework}, the hypothesis-to-code factor discovery loop begins by translating planned hypotheses into executable Python seed factors using daily OHLCV data.
Each candidate includes a financial hypothesis, executable code, and an explanation of the implementation logic.
During \textbf{validation and repair}, static inspection first checks that the implementation follows the data and code contract.
Tri-alignment then checks whether the hypothesis and implementation logic agree with the executable code and whether both code and explanation are financially coherent.
Detected inconsistencies trigger targeted repair and revalidation, keeping the research record consistent with the implemented mechanism for subsequent attribution and feedback.
Candidates that pass these checks are executed to assess output validity and screen for future-information leakage.
Recoverable problems trigger targeted repair followed by revalidation; persistent failures are discarded before empirical evaluation.

During \textbf{empirical evaluation and selection}, valid candidates are assessed for cross-sectional predictive quality using IC and RankIC, with factor direction fixed from training data.
ICIR and RankICIR summarize predictive stability.
A hard rolling out-of-sample (OOS) filter screens candidates for reliable performance across successive evaluation windows.
Normal selection uses train-only evidence to maintain a broad parent pool for further evolution.
A complexity discount adjusts its predictive score to favor simpler implementations with comparable predictive quality.
Elite selection additionally uses validation-period evidence in train--validation scoring and applies stricter quality and rolling-OOS requirements.
Selected elites are preserved across generations in an archive that supports feedback construction and final portfolio evaluation.
Held-out test results do not inform research selection or feedback.
The scoring and selection rules are detailed in Appendix~\ref{app:factor_selection_details}.
Section~\ref{sec:experimental_settings} specifies the final portfolio reporting protocol.

Retained factors serve as parents for \textbf{code-based evolution}.
Mutation introduces variants of a parent mechanism, and crossover combines parent hypotheses into a joint mechanism.
Refinement simplifies an implementation while preserving its core intent.
Generated children return to validation and empirical evaluation, closing the generation-level loop.
\textbf{Diversity control} removes duplicate implementations and filters candidates that are highly correlated with existing elite factors before selection.
Within each cycle, periodic \textbf{novelty injection} uses the cycle theme and recent success/failure feedback to recompose research agents and generate fresh seeds beyond variants of the current parents.
These seeds enter the parent pool after redundancy screening, and their descendants undergo the same validation, evaluation, and selection process as other evolved candidates.
Details are in Appendix~\ref{app:hypothesis_to_code}.

\vspace{-0.2cm}
\subsection{Feedback and Memory Update}
\label{sec:feedback_memory_update}
\vspace{-0.2cm}

As shown in Figure~\ref{fig:xalpha_framework}, feedback consolidation turns factor-evaluation outcomes into reusable research experience.
For \textbf{factor-to-archetype attribution}, code-validated candidates are mapped to mechanism families and \emph{Research Archetypes} under the shared taxonomy used by RMA.
Attribution follows the implemented mechanism, which may change during repair or evolution.
During \textbf{feedback construction}, elite-selected factors contribute GOOD records that capture supported mechanisms and conditions for reuse.
Normal-gate failures contribute BAD records that explain failed assumptions and provide avoidance or repair guidance.
Recent GOOD/BAD summaries guide fresh-seed generation during within-cycle novelty injection.
For \textbf{memory update}, only elite-factor hypotheses are written to archetype-specific memory under their attributed \emph{Research Archetypes}.
Cycle-level findings are consolidated into persistent summaries within the explored research scope, retaining the remaining information gaps.
Subsequent routing uses this experience to identify the next information gap and revise the research direction and search space, enabling memory-driven self-evolution across cycles.
Attribution and feedback-construction details are provided in Appendix~\ref{app:feedback_memory_update}.

\setlength{\textfloatsep}{10pt}
\begin{table*}[t]
\centering
\scriptsize
\setlength{\tabcolsep}{3.2pt}
\renewcommand{\arraystretch}{1.06}
\resizebox{\textwidth}{!}{
\begin{tabular}{llccccccccc}
\toprule
\rowcolor{xalphaHeader}
\multicolumn{2}{c}{\textbf{Models}} &
\multirow{2}{*}{\textbf{Dataset}} &
\multirow{2}{*}{\textbf{Target}} &
\multicolumn{7}{c}{\textbf{Evaluation Metrics}} \\
\cmidrule(lr){1-2} \cmidrule(lr){5-11}
\rowcolor{xalphaHeader}
\textbf{Type} & \textbf{Model} & & &
\textbf{IC} & \textbf{RankIC} & \textbf{ICIR} & \textbf{RankICIR} & \textbf{AR} & \textbf{AER} & \textbf{IR} \\
\midrule

\multirow{4}{*}{Machine Learning}
 & LightGBM & \multirow{4}{*}{CSI300} & \multirow{4}{*}{10-day} & 0.0213 & 0.0291 & \secondcell{0.2180} & \secondcell{0.2454} & 0.0683 & \thirdcell{0.0612} & \secondcell{0.9003$\pm$0.2259} \\
 & AdaBoost &  &  & 0.0161 & 0.0191 & \thirdcell{0.2051} & 0.2054 & 0.0536 & 0.0457 & 0.6714$\pm$0.1124 \\
 & XGBoost &  &  & 0.0161 & 0.0162 & 0.1942 & 0.1894 & 0.0564 & 0.0475 & 0.7167$\pm$0.0573 \\
 & Random Forest &  &  & 0.0117 & 0.0141 & 0.1391 & 0.1680 & 0.0553 & 0.0467 & 0.7010$\pm$0.0414 \\
\midrule

\multirow{4}{*}{Deep Learning}
 & Transformer & \multirow{4}{*}{CSI300} & \multirow{4}{*}{10-day} & 0.0079 & 0.0150 & 0.0749 & 0.1365 & 0.0473 & 0.0383 & 0.5459$\pm$0.3059 \\
 & LSTM &  &  & 0.0112 & 0.0240 & 0.1308 & 0.2115 & 0.0517 & 0.0414 & 0.5261$\pm$0.0956 \\
 & MLP &  &  & 0.0156 & 0.0237 & 0.1415 & 0.2043 & 0.0419 & 0.0331 & 0.4305$\pm$0.2849 \\
 & CNN &  &  & 0.0152 & 0.0226 & 0.1380 & 0.1995 & 0.0481 & 0.0381 & 0.4886$\pm$0.1431 \\
\midrule

\multirow{4}{*}{Alpha Methods}
 & CogAlpha & \multirow{4}{*}{CSI300} & \multirow{4}{*}{10-day} & \secondcell{0.0297} & \secondcell{0.0373} & 0.1796 & 0.2141 & 0.0649 & 0.0562 & 0.6088$\pm$0.4550 \\
 & R\&D-Agent(Q) &  &  & \thirdcell{0.0261} & \thirdcell{0.0334} & 0.1801 & \thirdcell{0.2361} & \secondcell{0.0756} & \secondcell{0.0669} & \thirdcell{0.7790$\pm$0.2437} \\
 & AlphaAgent &  &  & 0.0198 & 0.0233 & 0.1389 & 0.1616 & 0.0534 & 0.0459 & 0.5222$\pm$0.0592 \\
 & Alpha360 &  &  & 0.0129 & 0.0099 & 0.0855 & 0.0636 & 0.0317 & 0.0299 & 0.2793$\pm$0.0000 \\
\midrule

\multirow{4}{*}{Single LLM}
 & DeepSeek-v3.2 & \multirow{4}{*}{CSI300} & \multirow{4}{*}{10-day} & 0.0220 & 0.0225 & 0.1674 & 0.1642 & \thirdcell{0.0684} & 0.0599 & 0.7197$\pm$0.4235 \\
 & GLM-5.2 &  &  & 0.0132 & 0.0136 & 0.1056 & 0.1039 & 0.0508 & 0.0441 & 0.5262$\pm$0.2238 \\
 & GPT-OSS-120B &  &  & 0.0243 & 0.0303 & 0.1583 & 0.2004 & 0.0600 & 0.0518 & 0.5476$\pm$0.3911 \\
 & Kimi-K2.7-Code &  &  & 0.0240 & 0.0257 & 0.1891 & 0.1918 & 0.0636 & 0.0565 & 0.6336$\pm$0.1077 \\
\midrule

\rowcolor{xalphaOursRow}
Ours
& \method{} & CSI300 & 10-day
& \bestcell{0.0448}
& \bestcell{0.0590}
& \bestcell{0.2602}
& \bestcell{0.3379}
& \bestcell{0.1379}
& \bestcell{0.1291}
& \bestcell{1.3472$\pm$0.3105} \\
\bottomrule
\end{tabular}
}
\vspace{-0.3cm}
\caption{ CSI300 10-day rolling comparison.
Metrics are averaged over all five recorded seeds or factor groups; IR is reported as mean $\pm$ standard deviation.
Shading ranks the displayed methods only, with the best mean bolded and the second-best underlined.
Complete results (mean $\pm$ SD) and individual AER values are in Appendix~\ref{app:baseline_results}.
}
\label{tab:csi300_baselines}
\vspace{-0.5cm}
\end{table*}

\vspace{-0.2cm}
\section{Experiments}
\label{sec:experiments}
\vspace{-0.2cm}

\subsection{Experimental Settings}
\label{sec:experimental_settings}
\vspace{-0.2cm}

\textbf{Datasets and prediction task.}
We evaluate \method{} on five equity indices across three different markets and two prediction horizons: CSI300 at 10 and 30 days, and CSI500, CSI1000, SP500, and HSI90 at 10 days.
The main experiments focus on CSI300 with a 10-day prediction horizon.
All settings use Qlib-compatible daily adjusted OHLCV data~\citep{yang2020qlib}.
For \(h\in\{10,30\}\), the target is the future adjusted open-to-open return from the next trading day's open: \( y_{i,t}^{(h)} = \frac{O^{\mathrm{adj}}_{i,t+h+1}}{O^{\mathrm{adj}}_{i,t+1}} - 1, \) where \(O^{\mathrm{adj}}_{i,t}\) denotes the adjusted open price.
We use 2011--2020 for training, 2021 for validation, and 2022--2025 for testing.
Training and validation are used for discovery and selection; the test period is reserved for final reporting.

% \vspace{-0.2cm}
\textbf{Metrics.}
We measure predictive quality using the information coefficient (\textbf{IC}) and rank information coefficient (\textbf{RankIC}).
Their corresponding mean-to-standard-deviation ratios, \textbf{ICIR} and \textbf{RankICIR}, measure temporal stability.
Held-out portfolio diagnostics include annualized return (\textbf{AR}), annualized excess return (\textbf{AER}), and information ratio (\textbf{IR}).
Factor direction is fixed using training data.
Additional definitions and diagnostics are provided in Appendix~\ref{app:evaluation_details}.

\textbf{System and evaluation configuration.}
Unless otherwise specified, \method{} uses \texttt{gpt-oss-120b}~\citep{openai2025gptoss}.
By default, each cycle targets 64 initial factors and maintains up to 80 parents over 10 generations.
In the main comparison, all predictive and factor-based methods use expanding-history rolling fitting, refitted every 128 trading days for each non-overlapping prediction window, with an \(h+1\)-day embargo.
For portfolio reporting, elite factors are ranked by \emph{library score}, a train--validation predictive score combining normalized RankIC and IC.
We retain up to 40 factors with pairwise absolute correlation below $0.60$ and combine them using Ridge regression~\citep{hoerl1970ridge} with $\alpha=10.0$.
Appendix~\ref{app:experimental_protocol} provides preprocessing, backtesting, and runtime settings; Appendix~\ref{app:factor_selection_details} details factor ranking.

\vspace{-0.2cm}
\subsection{Comparison with Baselines}
\label{sec:comparison_baselines}
\vspace{-0.2cm}

Table~\ref{tab:csi300_baselines} compares \method{} with selected machine-learning (ML), deep-learning (DL), alpha-method, and single-LLM baselines under the common CSI300 10-day rolling setting.
The full comparison of 23 methods appears in Appendix~\ref{app:baseline_results}.
ML and DL baselines are evaluated with five different random seeds on the same OHLCV feature panel.
The alpha-method group includes the fixed Alpha360 library~\citep{qlib-alpha360} and automated mining systems~\citep{tang2025alphaagent,li2025rdagent,liu2026cognitive}.
Single-LLM baselines, generative alpha methods, and \method{} each generate 1,500 valid candidate factors.
We select five 40-factor groups in descending train--validation library-score order, excluding factors already chosen for earlier groups.
Within each group, a candidate is retained only when its absolute correlation with every selected factor is below $0.60$.
The factors in each group are combined using Ridge regression and evaluated under the rolling protocol in Section~\ref{sec:experimental_settings}.
For every method, we average all seven metrics over its five seed- or group-level outcomes.

\method{} achieves the highest mean on all seven metrics in the main table.
Its held-out RankIC is 0.0590, compared with 0.0373 for CogAlpha, the strongest displayed baseline on this metric.
In portfolio evaluation, its AER reaches 0.1291, compared with 0.0669 for R\&D-Agent(Q).
Its IR is $1.3472\pm0.3105$, compared with $0.9003\pm0.2259$ for LightGBM.
These results support the predictive quality and portfolio usefulness of the discovered factor groups under the shared evaluation protocol.
We also evaluate \method{} with different factor-group sizes and correlation thresholds; the combination results are reported in Appendix~\ref{app:xalpha_selection_sensitivity}.
Appendix~\ref{app:baseline_results} provides the complete rolling and static comparisons, together with individual AER values and cumulative-return curves.

\vspace{-0.2cm}
\subsection{Market and Horizon Generalization}
\label{sec:cross_market_generalization}
\vspace{-0.2cm}

\begin{figure}[t]
\centering
\includegraphics[width=\columnwidth]{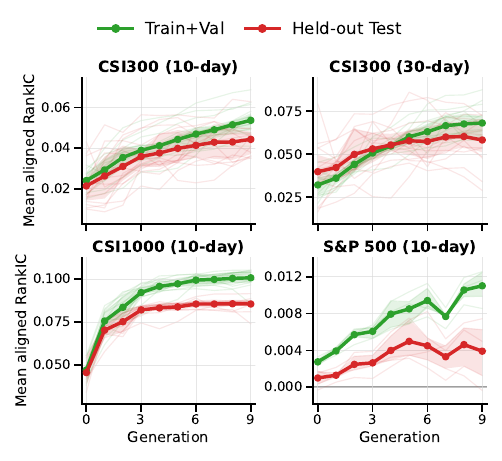}
\vspace{-0.9cm}
\caption{Generation-level RankIC in 4 settings.
Thin lines show individual cycles; bold curves and shading indicate the mean and interquartile range.}
\vspace{-0.2cm}
\label{fig:cross_market_generation_rankic}
\end{figure}

Figure~\ref{fig:cross_market_generation_rankic} examines whether generational improvement extends beyond CSI300 at the 10-day horizon, using four settings spanning horizon, equity universe, and market.
For each generation, we average RankIC over the top 8 elite factors within each cycle, then average the cycle-level means across 15 cycles.
Approximately 100 elite factors contribute to each generation-level estimate.
Factor directions are fixed from training data (Appendix~\ref{app:direction_alignment}).
Train--validation curves summarize evidence available during research, whereas held-out curves summarize test-period factor quality.
The figure assesses how generation-level factor quality generalizes across markets and horizons.

From the initial to the final generation, mean held-out RankIC increases +106.9\% on CSI300 at 10 days and +46.3\% at 30 days.
At the 10-day horizon, it rises +114.0\% on CSI500 and +87.2\% on CSI1000.
These trajectories show that the improvement is not confined to one A-share universe or one prediction horizon.
HSI90 and SP500 also improve, +181.4\% and +295.1\%, respectively.
These substantial relative gains across all evaluated markets support the cross-market generalization of \method{}'s factor-discovery capability.
Appendix Figure~\ref{fig:cross_market_generation_rankic_full} presents all six settings.

\vspace{-0.2cm}
\subsection{Evaluating the Core Memory Mechanisms}
\label{sec:core_memory_evaluation}
\vspace{-0.2cm}

\textbf{Report-to-Memory Abstraction.}
We test whether RMA improves research-input quality by comparing factor generation from report-grounded research paths with \emph{direct report} input.
Both conditions use 2,120 reports and the same factor-discovery pipeline.
Table~\ref{tab:rma_predictive_quality} shows relative gains in mean held-out metrics across all accepted factors.
RMA improves all four metrics, with RankIC and RankICIR increasing by 6.40\% and 29.76\%, respectively.
When these factors are combined, RMA also improves RankIC and RankICIR in all 12 factor-count/correlation configurations.
These results support the effectiveness of RMA in providing more actionable research input and improving the predictive quality of generated factors.
Appendix Tables~\ref{tab:rma_predictive_quality_full} and~\ref{tab:rma_combination_grid} provide the complete single-factor and combination results.

\textbf{Memory-driven research routing.}
We compare memory-driven routing with fixed and random routing under the same setting.
Fixed routing retains the same theme and archetype bundle across cycles.
We run five independent experiments per condition, each sampling 20 factors from the initial generation of each of five cycles (100 factors total).
Table~\ref{tab:routing_diversity} shows a 35.6\% reduction in mean absolute cross-cycle factor correlation under memory-driven routing.
Effective rank increases by 67.4\%.
Random routing yields intermediate diversity between fixed and memory-driven routing.
We further assess the quality of candidates selected by Train--Val RankIC.
Memory-driven routing achieves mean held-out RankIC of 0.02634 and 0.01843 for the top-10 and top-20 subsets, exceeding fixed routing by 27.9\% and 15.3\%, respectively (Appendix Table~\ref{tab:routing_selected_quality}).
Together, these results suggest that memory-driven routing reduces repeated exploration across cycles and broadens signal diversity, yielding a stronger upper tail of selectable high-quality factors.
Appendix~\ref{app:cross_cycle_factor_discovery} provides a feedback-to-routing trace, and Appendix Figure~\ref{fig:routing_heatmap_appendix} visualizes the diversity contrast.

\begin{table}[t]
\centering
\small
\setlength{\tabcolsep}{3pt}
\begin{tabular*}{\columnwidth}{@{\extracolsep{\fill}}lrrrr@{}}
\toprule
 & IC & RankIC & ICIR & RankICIR \\
\midrule
Gain & +46.28\% & +6.40\% & +137.09\% & +29.76\% \\
\bottomrule
\end{tabular*}
\vspace{-0.2cm}
\caption{Relative gains over \emph{direct report} input in mean held-out single-factor metrics on CSI300 (10-day).
}
\label{tab:rma_predictive_quality}
\vspace{-0.2cm}
\end{table}

\begin{table}[!htbp]
\centering
\small
\begin{tabular*}{\columnwidth}{@{\extracolsep{\fill}}lrrr@{}}
\toprule
Metric & Fixed & Random & Memory-driven \\
\midrule
Mean $|\mathrm{corr}|$ $\downarrow$ & $0.2106$ & $0.1980$ & $\mathbf{0.1357}$ \\
P90 $|\mathrm{corr}|$ $\downarrow$ & $0.5374$ & $0.5210$ & $\mathbf{0.3489}$ \\
Effective rank $\uparrow$ & $27.01$ & $29.84$ & $\mathbf{45.22}$ \\
\bottomrule
\end{tabular*}
\vspace{-0.3cm}
\caption{Cross-cycle signal diversity across five independent experiments.}
\label{tab:routing_diversity}
\vspace{-0.3cm}
\end{table}

\vspace{-0.2cm}
\subsection{Closed-Loop Process Evaluation}
\label{sec:closed_loop_evaluation}
\vspace{-0.2cm}

We next evaluate how the supporting operations of the closed-loop research process affect factor quality and the reliability of accumulated research experience.
Study-specific protocols and complete results appear in Appendix~\ref{app:ablation_protocols}.

\textbf{Feedback-guided discovery.}
We compare real discovery feedback with no feedback and randomly sampled feedback from the same candidate bank, generating 100 candidates per condition in each experiment.
We compare equally sized candidate subsets selected by train--validation score at $K\in\{5,10,20,50\}$ and also evaluate all accepted candidates.
Table~\ref{tab:discovery_feedback_quality} presents the top-5 comparison, where real feedback achieves the highest mean held-out RankIC and RankICIR across the two experiments.
This advantage holds across the other selection sizes and the all-valid pools.
These results support the effectiveness of historical success and failure feedback in guiding factor generation and improving the predictive quality of subsequent factors.
Appendix~\ref{app:feedback_k_sensitivity} provides the complete results.

\textbf{Reliable research records through tri-alignment.}
We randomly sample 167 factor records from 1,500 candidates for a paired comparison before and after tri-alignment repair.
Manual review assesses explanation--code agreement and the financial soundness of both code and explanation.
The proportion passing all three checks increases from 12.57\% to 89.82\% after repair.
These results support improved consistency among hypotheses, factor code, and financial explanations, strengthening the quality of research-memory entries.
Appendix~\ref{app:ablation_protocols} provides the repair-audit details.

\begin{table}[t]
\centering
\small
\begin{tabular*}{\columnwidth}{@{\extracolsep{\fill}}lrr@{}}
\toprule
Feedback & RankIC & RankICIR \\
\midrule
None & 0.008456 & 0.063165 \\
Random & 0.006767 & 0.057163 \\
Real & \textbf{0.042562} & \textbf{0.261827} \\
\bottomrule
\end{tabular*}
\vspace{-0.3cm}
\caption{Held-out quality of top-5 factors selected by train--validation score, averaged across two experiments.}
\label{tab:discovery_feedback_quality}
\vspace{-0.3cm}
\end{table}

\textbf{Diversity refresh through novelty injection.}
We select 10 novelty-injection events and compare the parent pool immediately before and after injection.
Mean absolute parent-pool correlation decreases in all 10 events, with a 5.90\% reduction in the average across events.
The event-averaged raw effective rank increases by 12.87\%.
Effective rank normalized by factor count increases in 8 events, and its event-averaged value rises by 2.68\%.
These results support the effectiveness of novelty injection in reducing factor redundancy during within-cycle evolution.
Appendix Table~\ref{tab:novelty_injection_events} reports the complete event-level results.

\textbf{Rolling OOS filtering.}
We compare factor selection with and without a hard filter based on rolling OOS performance, using the same evaluated candidate bank.
The filter checks temporal robustness within the train--validation period.
At top-200 in the normal pool, hard filtering improves mean held-out RankIC and RankICIR by 14.56\% and 17.85\%, respectively, over selection without rolling OOS.
These results support the effectiveness of rolling OOS filtering in identifying more reliable candidates and improving selection quality.
Appendix Table~\ref{tab:rolling_oos_selection} compares selection with and without rolling OOS filtering in both pools.

\textbf{Complexity discounting.}
We compare the four complexity-discount options in \method{} by selecting 40 factors from the same fixed RMA candidate pool.
All four improve held-out predictive quality and reduce mean factor complexity compared with selection without discounting.
RankIC improves by 5.07--7.98\% and RankICIR by 2.77--5.32\% across the four functions.
Mean complexity falls by 7.31--12.69\%.
These results suggest that appropriate complexity control can improve factor quality.
Appendix Table~\ref{tab:complexity_quality} reports all four functions and the no-discount control.

% \vspace{-0.2cm}
\subsection{Analysis of Generated Factors}
\label{sec:component_analysis}
\vspace{-0.2cm}

\textbf{Within-theme signal diversity.}
Once a research theme is fixed, we examine whether factors generated within that search space retain signal diversity.
We analyze 100 factors from one volatility-regime breakout cycle.
Across the 4,950 factor pairs, the mean absolute signal correlation is 0.290.
Absolute correlation is below 0.3 for 56.6\% of pairs.
The 100-factor correlation matrix has an effective rank of 20.73.
These results show variation in factor signals within a shared theme.
Appendix Figure~\ref{fig:within_theme_factor_corr_heatmap} provides the correlation heatmap and full statistics.

\textbf{Hypothesis-to-code case study.}
The following pseudocode summarizes an elite CSI300 factor based on regime-dependent overshoot pressure.

\begin{lstlisting}[language={},basicstyle=\footnotesize\ttfamily\color{black},numbers=none,aboveskip=4pt,belowskip=4pt,xleftmargin=0pt,framexleftmargin=0pt]
ret = FILLNA(daily_return, 0)
cutoff = WHERE(high_vol, 0.90, 0.95)
overshoot = close < cutoff * MA20
pressure = SUM20(overshoot * abs(ret))
fast = EWMA(pressure, span=5)
slow = EWMA(pressure, span=15)
signal = WHERE(high_vol, fast, slow)
factor = signal - HIST_Q75(signal)
\end{lstlisting}

Here, \texttt{MA20} and \texttt{SUM20} denote trailing 20-day mean and sum operations, with missing returns set to zero.
High volatility is identified when 20-day return volatility exceeds its 60-day rolling median.
The two fixed-span EWMAs maintain separate histories, and \texttt{WHERE} selects their outputs at each date.
Choosing the fast output in high-volatility states expresses the hypothesis of faster reversal after downside overshoot.
Subtracting \texttt{HIST\_Q75}, the expanding 75th percentile, centers the signal relative to historical pressure levels.
Appendix~\ref{app:representative_factor_code} provides the implementation and evaluation results.

\vspace{-0.2cm}
\section{Conclusion}
\label{sec:conclusion}
\vspace{-0.2cm}

We present \method{}, a memory-driven AI Quant Researcher for continuous hypothesis-to-code alpha discovery.
By combining report-grounded knowledge with persistent discovery feedback, \method{} revises research directions and search spaces across cycles, enabling memory-driven self-evolution.
Experiments across five different equity indices and three distinct markets evaluate this framework, with comprehensive comparisons on CSI300 supporting improvements in factor quality and signal diversity.
This work provides a foundation for cumulative, experience-driven quantitative research, supporting the development of AI researchers that learn from prior investigations and adapt their exploration over time.

\section*{Limitations}

\method{} is intended for academic research and does not provide financial advice.
Factor discovery is restricted to daily OHLCV data, limiting the financial mechanisms that can be investigated.
Portfolio evaluation uses historical Qlib backtests with fixed trading-cost assumptions, which only approximate live trading conditions.
Additionally, the stochastic nature of LLM outputs makes exact reproduction of generated factors difficult.
Discovery quality also depends on the chosen language model and evaluation configuration.

\section*{Ethical Considerations}

The market datasets used in this study were obtained from public sources through Qlib~\citep{yang2020qlib} and Yahoo Finance~\citep{Aroussi2024}.
Users should follow the applicable data-use terms and independently verify generated factors and code before deployment.
Any practical use requires an assessment of the associated financial risks.

AI assistance in manuscript preparation was limited to grammar correction and refinement of logical clarity.
All edits were reviewed by the authors, who take responsibility for the final manuscript.

\bibliography{xalpha_refs}

%%%%%%%%%%%%%%%%%%%%%%%%%%%%%%%%%%%%%%%%%%%%%%%%%%%%%%%%%%%%

\clearpage
\onecolumn
\appendix

\section{Appendix Organization}
\label{app:organization}

The appendix follows the research workflow before presenting supporting experiments.
Appendix~\ref{app:runtime_pseudocode} summarizes the overall runtime loop.
Appendix~\ref{app:rma_details} describes report-to-memory abstraction, and Appendix~\ref{app:archetype_taxonomy} defines the shared research taxonomy.
Appendix~\ref{app:memory_routing} explains research planning and routing, with examples of routing decisions and cross-cycle self-evolution.
Appendix~\ref{app:hypothesis_to_code} details the hypothesis-to-code factor discovery loop.
Appendix~\ref{app:factor_selection_details} covers factor evaluation and selection.
Appendix~\ref{app:feedback_memory_update} explains how factor attribution supports feedback consolidation and memory update.

Appendix~\ref{app:evaluation_details} provides metric definitions and scoring rules.
Appendix~\ref{app:experimental_protocol} specifies the experimental protocol.
Appendix~\ref{app:baseline_results} reports complete rolling and static baseline comparisons and factor-selection sensitivity results.
Appendix~\ref{app:ablation_protocols} provides component-study protocols and detailed results.

Appendix~\ref{app:case_studies} presents generated-factor diversity analysis and factor-code case studies.
Appendix~\ref{app:primitive_features} defines the primitive features, and Appendix~\ref{app:prompt_excerpts} presents selected key prompt excerpts.

\section{Overall Runtime Pseudocode}
\label{app:runtime_pseudocode}

The pseudocode summarizes the research loop, with selection and memory updates guided by training and validation evidence.

\begin{xpromptlisting}{Research loop of \method{}}
Input:
  report-grounded research memory and shared taxonomy
  persistent cycle-feedback and archetype-specific hypothesis memory
  daily OHLCV panels, primitive features, and prediction horizon h
  train, validation, and held-out test periods
  num_cycles, cycle_len, pool_size, fresh_seed_interval

Initialize:
  base_pool <- raw and primitive baseline factors
  reporting_candidates <- empty
  retain existing research and feedback memories

EvaluateCandidates(candidates):
  valid <- validate_and_repair_code_and_factor_series(candidates)
  valid <- check_and_repair_tri_alignment(valid)
  attributed <- assign_mechanism_family_and_archetype(valid, taxonomy)
  evaluated <- compute_predictive_metrics(attributed, train, validation,
      direction_fixed_from_train=True)
  evaluated <- compute_rolling_OOS_diagnostics(evaluated)
  record_test_diagnostics(evaluated, direction_fixed_from_train)
  return evaluated

SelectCandidates(evaluated):
  normal_pass, normal_fail <- apply_normal_gate(evaluated, split=train)
  normal_pool <- select_parent_pool(normal_pass, capacity=pool_size,
      score=train_alpha_score * complexity_discount)
  elite_pass <- apply_stricter_elite_gate(evaluated, split=train+validation)
  current_elites <- select_elite_pool(elite_pass,
      score=train_validation_alpha_score)
  # Both gates apply rolling OOS hard filtering.
  return normal_pool, current_elites, normal_fail

For each research cycle c = 1 ... num_cycles:
  normal_parent_pool <- empty
  elite_archive <- empty
  novelty_feedback <- empty
  cycle_feedback <- empty

  # Cycle-level research planning and routing.
  cycle_theme <- select_next_theme(recent_cycle_feedback_memory)
  families <- select_primary_and_supporting_families(cycle_theme)
  archetypes <- select_relevant_archetypes(families, cycle_theme)
  paths <- retrieve_report_grounded_paths(archetypes, cycle_theme)
  active_agents <- compose_research_agents(archetypes, paths)
  hypothesis_pool <- construct_hypothesis_pool(active_agents)
  cycle_scope <- archetypes

  # Generation 0: seed discovery.
  subsets <- plan_hypothesis_subsets(hypothesis_pool)
  seeds <- generate_factor_code(active_agents, subsets)
  evaluated_seeds <- EvaluateCandidates(seeds)
  normal_seeds, seed_elites, _ <- SelectCandidates(evaluated_seeds)
  normal_parent_pool <- select_parent_pool(normal_seeds + base_pool)
  elite_archive <- seed_elites
  update_archetype_hypothesis_memory(elite_archive)

  # Generations 1 ... cycle_len - 1; finalization is a separate step.
  For generation g = 1 ... cycle_len - 1:
    parents <- deduplicate(elite_archive + normal_parent_pool)

    # Injection precedes evolution at G3 and G7 in the 10-generation setting.
    If (g + 1) is divisible by fresh_seed_interval:
      novelty_context <- summarize_GOOD_BAD(novelty_feedback)
      active_agents <- recompose_agents_within_theme(cycle_theme)
      cycle_scope <- union(cycle_scope, active_agents.archetypes)
      fresh_hypotheses <- plan_fresh_hypotheses(active_agents,
          archetype_hypothesis_memory, elite_archive)
      fresh_seeds <- generate_factor_code(active_agents,
          fresh_hypotheses, novelty_context)
      fresh_evaluated <- EvaluateCandidates(fresh_seeds)
      fresh_pass <- screen_redundancy_and_novelty_gate(fresh_evaluated,
          anchors=parents)
      parents <- deduplicate(parents + fresh_pass)
      clear(novelty_feedback)  # Retain cycle-level feedback.

    children <- evolve_factor_code(parents,
        operators={mutation, crossover, refinement})
    evaluated <- EvaluateCandidates(children)
    evaluated <- screen_duplicates_and_signal_correlation(evaluated,
        anchors=elite_archive)
    normal_parent_pool, current_elites, normal_fail <- SelectCandidates(evaluated)
    elite_archive <- merge_and_select_archive(elite_archive, current_elites)
    update_archetype_hypothesis_memory(elite_archive)  # Elite hypotheses only.
    records <- collect_GOOD_BAD(current_elites, normal_fail,
        evidence=train_and_validation_metrics)
    update_feedback_buffers(novelty_feedback, cycle_feedback, records)

  # Cycle-level feedback consolidation and memory update.
  reporting_candidates <- merge_elite_records(reporting_candidates, elite_archive)
  summary <- consolidate_cycle_feedback(cycle_theme, cycle_scope,
      elite_archive, cycle_feedback, evidence=train_and_validation_metrics)
  persist_cycle_feedback_memory(summary)

# Post-run portfolio evaluation: factor membership is fixed before testing.
groups, portfolio_results <- evaluate_reporting_portfolios(
    reporting_candidates, selection_evidence=train+validation)

Output:
  archived elite factors and generation-level diagnostics
  updated archetype-specific hypotheses and persistent cycle feedback
  fixed reporting groups and portfolio evaluation results
\end{xpromptlisting}

\section{Report-to-Memory Abstraction Details}
\label{app:rma_details}

The pipeline and examples below expand the RMA overview in Figure~\ref{fig:rma_abc_taxonomy}.

\subsection{Pipeline Implementation}

\paragraph{Document ingestion and A-layer filtering.}
PDF reports are converted to Markdown and split into evidence units.
The A-layer checks whether each unit contains a self-contained mechanism that can be studied using daily OHLCV, including stable proxies derived from these data.
Mixed passages are retained when they contain at least one eligible mechanism.
Failed or invalid screening calls use a conservative KEEP fallback to reduce false negatives.
Data compatibility is screened again during research-path extraction.

\paragraph{B-layer extraction and classification.}
The B-layer extracts specific research paths from screened chunks or report segments.
Each accepted path must be expressible with daily OHLCV and is assigned to one B-layer mechanism family.
A single evidence unit may yield multiple paths, potentially belonging to different families.
Accepted paths retain report evidence and undergo output validation before archetype assignment.

\paragraph{C-layer routing and memory construction.}
Each accepted B-layer path is routed to one compatible C-layer archetype within its B-family.
The LLM returns a C-layer identifier, which is checked against the allowed B-family.
Valid paths are grouped by archetype for embedding-based similarity de-duplication.
A path joins a similarity group only if it meets the similarity threshold against every existing member.
The path with the highest mean within-group similarity is retained as the representative.
Subsequent memory curation consolidates closely related paths into more abstract research descriptions.
The resulting records associate each archetype with its mechanism description and curated research paths.

\paragraph{Shared research taxonomy.}
The predefined B/C taxonomy is shared with factor-to-archetype attribution (Appendix~\ref{app:feedback_memory_update}).

\subsection{Report-Based Examples}
\label{app:rma_trace_case}

The examples show LLM-call excerpts within the pipeline and a curated memory entry after consolidation.
Excerpts preserve the original wording with typographic normalization; \texttt{...} marks omissions.

\paragraph{A/B/C continuation example.}
The example follows bullish moving-average alignment from a report on pullback and oversold rebound through A-layer screening, B1 extraction, and C102 routing.
The A-layer screens the full passage, which also discusses rebound mechanisms.

\begin{xpromptlisting}{A-layer screening: prompt and report evidence excerpts}
[system]
You are applying Layer A of the OHLCV report taxonomy.
...
Layer A question:
Can this report fragment enter the daily Level-1 OHLCV research workflow at all?
...
Important:
- Follow Layer A exactly; do not invent a separate screening standard.
- This is an eligibility decision, not a mechanism-family classification task.
- Focus on the core mechanism, not on surface vocabulary alone.
- Do not summarize the whole chunk.
- Do not propose factor implementations.

[user]
CHUNK (Markdown):

# Moving-Average Pullback Oversold Rebound Stock-Selection Strategy
...
Moving-average trends at different frequencies reflect market trend changes and give rise to various trend-following strategies. The trajectory of a moving average represents the market's transaction cost and chip distribution over the corresponding period, and to some extent reflects investor sentiment. When the moving average shows a bullish (ascending) arrangement, market cost gradually rises and investors' greed to chase the rally intensifies with the trend.
...
\end{xpromptlisting}

\begin{xpromptlisting}{A-layer response: OHLCV eligibility}
{
  "decision": "KEEP",
  "why": "the core mechanism (moving-average pullback and oversold rebound) can be expressed using daily price data alone"
}
\end{xpromptlisting}

\begin{xpromptlisting}{B-layer extraction: prompt excerpts}
[system]
Mine daily-OHLCV research paths from ONE report chunk, then assign each path to one Layer-B family.

Input assumptions:
- The chunk already passed Layer A.
- Available data is daily open/high/low/close/volume only.
- Do not output formulas, indicators, backtests, portfolios, risk rules, C-layer ids, or agent names.
...
Definition: Studies continuation, trend formation, momentum reinforcement, and momentum decay.
...
\end{xpromptlisting}

\begin{xpromptlisting}{B-layer response excerpt: Trend and Momentum}
{
  "status": "accepted",
  "research_path": "Analyze how concurrent bullish alignment of multiple daily moving averages (e.g., 20-day and 50-day) predicts sustained price momentum and trend continuation",
  "b_id": "B1",
  "b_layer_reason": "Studies trend formation and momentum reinforcement via bullish moving-average alignment, fitting B1's trend and momentum category",
  ...
  "evidence_quote": "When the moving average shows a bullish (ascending) arrangement, market cost gradually rises and investors' greed to chase the rally intensifies",
  ...
}
\end{xpromptlisting}

\paragraph{C-layer common prompt and three B1 examples.}
The C-layer examples use the three \emph{Trend \& Momentum} archetypes defined in Appendix~\ref{app:archetype_taxonomy}.
In the quoted prompt, \emph{Layer-C agent} denotes a research archetype.

\begin{xpromptlisting}{C-layer shared prompt excerpts}
[system]
Route ONE research-path atom to exactly one Layer-C agent from the allowed catalog.
...
Available c_id options:
C101, C102, C103
...
Rules:
- Choose exactly one c_id from the provided Layer-C catalog.
...

Return ONLY valid JSON:
{
  "c_id": "<one c_id from Available c_id options>"
}
\end{xpromptlisting}

\begin{xpromptlisting}{C102: Trend Continuation}
Input excerpt:
Research Path: Analyze how concurrent bullish alignment of multiple daily moving averages (e.g., 20-day and 50-day) predicts sustained price momentum and trend continuation
...

Response:
{
  "c_id": "C102"
}
\end{xpromptlisting}

\begin{xpromptlisting}{Curated memory path: Trend Continuation}
Measure completed short/medium/long average ordering and normalized separation, time since a crossover and persistence of alignment; test continuation/decay beyond return and a single cross, treating SMA/EMA as defined alternatives.
\end{xpromptlisting}

\paragraph{Complementary archetype examples.}
Two complementary examples from other report passages illustrate the remaining archetypes: a short-term moving average crossing above a long-term average suggests early trend initiation, whereas declining moving-average slope and volume after a rally suggest momentum weakening.

\begin{xpromptlisting}{C101: Trend Initiation}
Input excerpt:
Research Path: Study daily short-term MA crossing above long-term MA as an early uptrend momentum signal, requiring 1-3 consecutive days of cross persistence and an open-close average higher than two days prior.
Keywords: short-term MA, long-term MA cross, early uptrend, 1-3 day persistence, open-close average rise
Supporting Logic: The pattern consists of three moving averages-short, medium, and long. | The short-term moving average crosses above the long-term moving average and can form within a single day or over 2-3 consecutive days. | The average of open and close on day T must exceed the same average on day T-2.

Response:
{
  "c_id": "C101"
}
\end{xpromptlisting}

\begin{xpromptlisting}{C103: Trend Exhaustion and Momentum Failure}
Input excerpt:
Research Path: Analyze daily trend slowdown after a modest market rally (<=10% rise) by measuring 20-day moving-average slope and volume contraction as proxies for weakening momentum
Keywords: modest rally, momentum decay, 20-day slope, volume contraction, cross-sectional reversal
Supporting Logic: Scenario 2 is triggered when the market rises <=10% and the 12-month average gain declines for two consecutive months, indicating weakening momentum. | In the month after such triggers, the high-momentum portfolio underperforms by 33.7% and suffers a 92.9% loss occurrence, showing a clear reversal effect. | A daily-level slowdown in price slope combined with shrinking volume should capture the same weakening momentum signal.

Response:
{
  "c_id": "C103"
}
\end{xpromptlisting}

\section{C-layer Research Archetype Taxonomy}
\label{app:archetype_taxonomy}

Table~\ref{tab:c-layer-archetypes} lists the 48 C-layer research archetypes in the shared research taxonomy.
Each archetype belongs to a B-layer mechanism family and groups research paths around a specific financial mechanism.
These labels link report-grounded knowledge with discovery feedback and guide knowledge retrieval during hypothesis planning.

\begingroup
\small
\setlength{\tabcolsep}{4pt}
\renewcommand{\arraystretch}{1.12}
\begin{longtable}{>{\raggedright\arraybackslash}p{0.15\linewidth}%
                  >{\raggedright\arraybackslash}p{0.30\linewidth}%
                  >{\raggedright\arraybackslash}p{0.47\linewidth}}
\toprule
\textbf{Archetype ID} & \textbf{Name} & \textbf{Role} \\
\midrule
\endfirsthead
\toprule
\textbf{Archetype ID} & \textbf{Name} & \textbf{Role} \\
\midrule
\endhead
\midrule
\multicolumn{3}{r}{\footnotesize Continued on next page} \\
\endfoot
\bottomrule
\caption{C-layer Research Archetype definitions.}
\label{tab:c-layer-archetypes}\\
\endlastfoot
\path{B1/C101} & Trend Initiation and Early Diffusion & Price escapes a local equilibrium and begins early directional diffusion. \\
\path{B1/C102} & Trend Continuation and Momentum Reinforcement & An existing trend becomes stronger and more persistent. \\
\path{B1/C103} & Trend Exhaustion and Momentum Failure & Continuation weakens and momentum begins to fail. \\
\path{B2/C201} & Short-Horizon Overreaction Reversal & A sharp local overreaction is followed by short-horizon repair. \\
\path{B2/C202} & Post-Capitulation Repair & Concentrated selling pressure becomes exhausted and price starts to recover. \\
\path{B2/C203} & Failed Breakout and Rejection Reversal & A directional attempt fails and price reverses after rejection. \\
\path{B3/C301} & Upper-Bound Breakout Hold & Price breaks resistance and successfully holds the new level. \\
\path{B3/C302} & Lower-Bound Breakdown Continuation & Price loses support and continues weakening. \\
\path{B3/C303} & Retest Confirmation and Relaunch & A post-breakout retest succeeds and the move resumes. \\
\path{B4/C401} & Body Strength and Close Location & Body size and close location capture local directional strength. \\
\path{B4/C402} & Shadow Pressure and Support & Upper and lower shadows reveal intraday rejection or absorption. \\
\path{B4/C403} & Multi-Bar Turning Pattern & Multiple bars jointly form a local turning or confirmation structure. \\
\path{B5/C501} & Gap Continuation and Overnight Diffusion & The overnight move continues to diffuse in the same direction. \\
\path{B5/C502} & Gap Fill and Overnight Overreaction & Overnight pricing overshoots and is corrected intraday. \\
\path{B5/C503} & Overnight-Intraday Role Switch & Overnight and intraday legs switch roles between weakness and recovery. \\
\path{B6/C601} & Volatility Compression Setup & Volatility contracts and creates a setup for later directional release. \\
\path{B6/C602} & Post-Expansion Volatility Repair & Volatility and price expand sharply and then cool down or repair. \\
\path{B6/C603} & Range Energy Build-Up and Release & The range structure builds and releases directional energy. \\
\path{B7/C701} & High-Volatility Regime Signal & The signal is primarily active in high-volatility or high-shock environments. \\
\path{B7/C702} & Low-Volatility Regime Signal & The signal is primarily active in low-noise or low-volatility environments. \\
\path{B7/C703} & Regime Transition Trigger & The transition between regimes becomes the predictive signal. \\
\path{B8/C801} & Downside Volatility Bias & Volatility responds more strongly to downside moves than to upside moves. \\
\path{B8/C802} & Tail-Risk Accumulation & Small fragility signals accumulate into tail-risk pressure over time. \\
\path{B8/C803} & Pre-Crash Fragility and Acceleration & Market fragility accelerates into instability. \\
\path{B9/C901} & Abnormal Volume Expansion Confirmation & A sudden rise in participation confirms a price-state change. \\
\path{B9/C902} & Volume Dry-Up and Consolidation & Participation shrinks and price enters consolidation or selling exhaustion. \\
\path{B9/C903} & Volume Structure Shift and Activity Repricing & The center of trading activity shifts before price structure reprices. \\
\path{B10/C1001} & Price-Volume Confirmation & Price direction is reinforced by aligned volume behavior. \\
\path{B10/C1002} & Price-Volume Divergence Reversal & Price direction and volume support diverge, weakening continuation quality. \\
\path{B10/C1003} & Volatility-Volume Resonance & Volatility expansion and volume expansion reinforce each other during risk release. \\
\path{B11/C1101} & Illiquidity Reversal & Low liquidity pushes price too far and increases reversal pressure. \\
\path{B11/C1102} & Liquidity-Stress Trend Distortion & Deteriorating liquidity amplifies or distorts an existing trend. \\
\path{B11/C1103} & Impact Decay and Liquidity Recovery & Price impact is absorbed, liquidity recovers, and price rebalances. \\
\path{B12/C1201} & Persistent Buying Pressure Proxy & A slow upward path with shallow pullbacks indicates persistent buying pressure. \\
\path{B12/C1202} & Persistent Selling Pressure Proxy & Persistent weakness and weak rebounds indicate selling pressure. \\
\path{B12/C1203} & Pressure Absorption and Reversal & Visible pressure is absorbed over time and the path reverses. \\
\path{B13/C1301} & Progressive Drawdown Accumulation & Drawdown accumulates through persistent weakness rather than a single crash. \\
\path{B13/C1302} & Post-Drawdown Recovery & Recovery quality after a deep drawdown contains predictive information. \\
\path{B13/C1303} & Failed Recovery and Second-Leg Down & An apparent recovery fails and weakness returns in a second leg. \\
\path{B14/C1401} & Lagged Price Response to Volume & Volume shifts occur first and price responds later. \\
\path{B14/C1402} & Post-Shock Delayed Drift & Price continues drifting after the initial shock rather than repricing immediately. \\
\path{B14/C1403} & Underreaction and Slow Adjustment & The market initially underreacts and adjusts gradually afterward. \\
\path{B15/C1501} & Crowding-Driven Continuation & Synchronized behavior reinforces an existing direction. \\
\path{B15/C1502} & Crowding Unwind Reversal & Crowded behavior unwinds and creates sharp reversal pressure. \\
\path{B15/C1503} & Herding-Amplified Fragility & Herding amplifies volatility, liquidity stress, and tail fragility. \\
\path{B16/C1601} & Market Phase Transition & Different market phases are governed by different dominant mechanisms. \\
\path{B16/C1602} & Compression-Expansion-Resolution Path & Predictive content comes from the ordered sequence of compression, expansion, and directional resolution. \\
\path{B16/C1603} & Fractal Consistency and Multi-Scale Complexity & Consistency or rupture across time scales is itself informative. \\
\end{longtable}
\endgroup

\section{Research Planning and Routing Details}
\label{app:memory_routing}

\subsection{Routing Procedure}

\paragraph{Research direction selection.}
In memory-driven cycles, the planner reads compact feedback summaries from the four most recent completed cycles together with the mechanism-family taxonomy.
Each summary describes the research question explored and the scope covered by the selected archetypes.
Positive and negative memory relate that coverage to the mechanisms supported and assumptions challenged by empirical evaluation.
The planner identifies an unresolved information gap and jointly proposes a cycle theme and mechanism-family profile to investigate it.

\paragraph{Search-space definition.}
The routing profile selects one primary mechanism family and either three or four supporting families from the shared taxonomy.
The primary family anchors the research question.
Each supporting family has an explicit role in addressing the same information gap, such as supplying a complementary mechanism or a validity condition.
The selected families determine the candidate archetype pool.

\paragraph{Theme-conditioned knowledge retrieval.}
The planner uses theme-specific research directions to retrieve semantically relevant report-grounded paths for selected \emph{Research Archetypes}.
Additional paths are sampled within each archetype to preserve variation.

\paragraph{Hypothesis planning.}
Each active agent constructs a hypothesis pool from its selected archetype and retrieved research paths.
An LLM-based subset planner selects hypotheses for seed-factor generation.
During within-cycle novelty injection, archetype-level memory supplements the pool with previously validated hypothesis ideas.
Feedback construction is detailed in Appendix~\ref{app:feedback_memory_update}.

\subsection{Routing Decision Example}
\label{app:cross_cycle_factor_discovery}
This case shows how accumulated cycle-level feedback both redirects exploration and defines the next search space.
The following excerpts summarize the routing decision.

\begin{xpromptlisting}{Memory-routing input: accumulated research experience}
Previously explored directions:
- hidden-liquidity decay under market stress;
- volatility transitions through range compression;
- post-spike reversal decay; and
- tail-pressure buildup during low-volume consolidation.

Reusable findings (GOOD memory):
- Condition liquidity and pressure mechanisms on the market regime.
- Preserve continuous price-volume and pressure information.
- Use responsive decay to represent persistence after compression or shocks.
- Treat low-volume structure as context rather than a standalone signal.

Recurring failures (BAD memory):
- Rare binary triggers and isolated single-day volume flags.
- Intraday proxies whose horizons do not match the prediction target.
- Stacks of individually weak components without marginal validation.
- Added complexity without a clear mechanism-level contribution.
\end{xpromptlisting}

\begin{xpromptlisting}{Memory-routing output: next-cycle research agenda}
Selected information gap:
How do volume dry-up, expansion, and re-accumulation interact with the buildup and
decay of directional pressure before a volatility-regime transition?

Cycle theme:
volume_structure_pressure_regime_gap

Primary B-layer:
B12 | Directional Pressure and Order Imbalance
supporting_function: primary_line

Supporting B-layers:
B9 | Volume Level and Structure
supporting_function: complement

B7 | Volatility Regime
supporting_function: contrast

B14 | Lagged Response and Delayed Adjustment
supporting_function: boundary_condition

B2 | Reversal and Mean Reversion
supporting_function: failure_avoidance

C-layer path-planning cues:
- Track pressure accumulation and decay across a dry-up--spike--re-accumulation path.
- Contrast persistent regime transitions with transient volatility spikes.
- Measure the delay between volume-pressure changes and subsequent price adjustment.
- Use rejected breakouts and rapid reversals as counterexamples to pressure triggers.
\end{xpromptlisting}

The router redirects research from isolated mechanisms toward their unresolved temporal interaction.
It defines the search space around directional pressure, with the supporting families providing volume sequence, regime discrimination, and transition-validity conditions.
The planner translates these roles into research paths, using avoidance rules to guide hypothesis construction.

\subsection{Cross-Cycle Research Self-Evolution}
\label{app:cross_cycle_self_evolution}
This CSI300 10-day case traces feedback-driven research redirection and the resulting memory update.

\paragraph{Prior research: cycles 6--9.}
The preceding research themes were:
\begin{itemize}
\item \textbf{Cycle 6:} \emph{Directional pressure and volatility-regime shifts}.
\item \textbf{Cycle 7:} \emph{Liquidity-driven downside regime transitions}.
\item \textbf{Cycle 8:} \emph{Pressure decay and asymmetric range compression}.
\item \textbf{Cycle 9:} \emph{Lagged pressure absorption under excess liquidity}.
\end{itemize}
Accumulated feedback supported liquidity-aware, volatility-conditioned signals and identified weak standalone shape proxies.
These findings motivated testing candlestick signals under explicit liquidity conditions.

\paragraph{Research redirection: cycle 10.}
\begin{itemize}
\item \textbf{Theme:} \emph{Candlestick--liquidity signals for regime transitions}.
\item \textbf{Research question:} Under what liquidity conditions can candlestick geometry distinguish early volatility-regime transition signals from shape noise?
\end{itemize}
The planner shifted to liquidity-conditioned candlestick signals.
One planned path examined whether long upper shadows with elevated daily volume signal volatility-regime transitions.

\paragraph{New memory after cycle 10.}
The cycle added:
\begin{itemize}
\item \textbf{Positive memory:} Retain volume- or volatility-conditioned signals and adaptive thresholds for isolating high-impact liquidity events.
\item \textbf{Negative memory:} Avoid raw shadow or volume measures as standalone predictors and unvalidated stacks of heterogeneous signals.
\end{itemize}
These findings guide subsequent research toward context-conditioned signals and away from unsupported shape-only constructions.

\section{Hypothesis-to-Code Factor Discovery}
\label{app:hypothesis_to_code}

\paragraph{Hypothesis-to-code generation.}
At cycle initialization, active research agents use planned hypothesis subsets and retrieved research paths to generate Python seed factors.
The generation context specifies the prediction horizon; implementations are restricted to approved daily OHLCV-derived inputs and transformations.
Each candidate provides a financial hypothesis (\texttt{hypothesis\_idea}), implementation logic (\texttt{logic}), and factor code.
The factor function receives one ticker's time-series slice with a \texttt{(date, ticker)} MultiIndex and returns an index-aligned \texttt{pd.Series}.
The returned series name must match the function name.
Available primitive features are listed in Appendix~\ref{app:primitive_features}.

\paragraph{Validation and repair.}
Static AST inspection checks the function interface, permitted operations, and implementation complexity.
It also screens for lookahead patterns, including negative shifts and backfilled signals.
Tri-alignment checks whether the explanatory text accurately describes the code after docstrings and comments are removed.
It separately reviews the financial coherence of the code and the explanation.
When only the explanatory text fails, repair preserves financially coherent code and revises the text.
Recoverable implementation errors trigger code repair using the failed checks and execution context.

Executable candidates produce ticker-wise factor series that are assembled into a panel.
The runtime checks index alignment and numerical validity, masking infinite or extreme values.
In the main setting, candidates are discarded when the global invalid-value ratio exceeds \texttt{nan\_thr}=0.30.
Invalid or uninformative cross-sectional dates are filtered separately.
Ticker-level truncation and future-noise perturbation tests provide additional leakage checks.
Repaired candidates repeat the validation pipeline; persistent failures are discarded before empirical scoring.
A recorded tri-alignment revision is presented in Appendix~\ref{app:case_studies}.

\paragraph{Empirical evaluation and selection.}
Validated factor series enter the evaluation and selection procedure in Appendix~\ref{app:factor_selection_details}.

\paragraph{Evolution and diversity control.}
Evolution uses retained parent code and hypotheses together with evaluation feedback.
Mutation creates a mechanism-level variant of a single parent factor.
Crossover combines two parent hypotheses into a joint mechanism implemented as code.
Refinement simplifies an implementation while preserving its core mechanism.
Score-aware parent sampling gives elite parents higher priority and retains normal parents for continued exploration.
Children return to validation and empirical evaluation before selection.
Deduplication and factor-correlation screening limit repeated implementations and redundant signals; the correlation criterion is defined in Appendix~\ref{app:redundancy_filtering}.

\textbf{Novelty injection} supplements parent-based evolution with fresh seeds for the current cycle theme.
\label{app:novelty_injection}
After generation-level selection, a short-lived buffer collects GOOD records from current-generation elites and BAD records from normal-gate failures.
It retains at most eight records of each type, with at most two per attributed agent and two per mechanism family.
Before injection, the LLM summarizes these records into reusable principles and avoidance or repair guidance, as described in Appendix~\ref{app:feedback_memory_update}.

Injection occurs when \((g+1)\) is divisible by four, where \(g\) is the zero-based generation index, excluding the final cycle step.
For a ten-generation cycle, this gives injection at generations 3 and 7.
Recomposed research agents generate fresh seeds using current-cycle hypothesis memory, the elite pool, and the summarized feedback.
The target batch size is half the initialization target: 32 fresh seeds for 64 initial factors.
Fresh seeds undergo the same validation and empirical evaluation as initial factors.

Fresh seeds are screened against leading previous elites and current parents with a correlation threshold of \(0.95\).
The correlation-screened batch then passes an internal gate: train-aligned winsorized ICIR and RankICIR must both reach the batch's 65th percentile.
Surviving seeds enter the current parent pool as normal-tier parents.
The feedback buffer is cleared after injection so that the next step uses newly accumulated experience.

\section{Factor Evaluation and Selection}
\label{app:factor_selection_details}

\paragraph{Evaluation inputs.}
Validated factor series are evaluated using direction-aligned IC and RankIC, with ICIR and RankICIR as stability diagnostics.
Direction is fixed from training RankIC and applied to the subsequent evaluation splits.
Let \(\alpha_s\) denote the split-level predictive score defined in Appendix~\ref{app:alpha_score}.
Metric definitions and rolling-OOS summaries are provided in Appendix~\ref{app:evaluation_details}.
The main setting uses rolling OOS as a hard filter, with its contribution to ranking scores disabled.
Before selection, child factors are deduplicated and dependent factors are removed.
Both static gates require aligned RankIC to meet the larger of the percentile threshold and fixed floor.

\paragraph{Normal selection.}
\label{app:normal_elite_selection}
Normal selection maintains candidates for further evolution using the train-only score
\[
s_{\mathrm{normal}}
=
\alpha_{\mathrm{train}}
\cdot
d_{\mathrm{complexity}},
\]
where \(d_{\mathrm{complexity}}\) is the implementation complexity discount (Appendix~\ref{app:complexity_discount}).
The static gate uses the 60th percentile for RankIC with a floor of \(0.005\), and requires a positive-day ratio of at least \(0.50\).
At least one secondary metric among aligned winsorized IC, ICIR, and RankICIR must reach its 50th percentile.
Warm-up replaces static gating with top-\(45\%\) score selection, keeping the rolling gate active.
After warm-up, gate-passed candidates proceed to parent-pool selection without score truncation.

\paragraph{Elite selection.}
Elite selection preserves stronger candidates across generations using train--validation predictive quality without a complexity discount:
\[
s_{\mathrm{elite}}
=
\alpha_{\mathrm{train+val}},
\]
The static gate uses the 80th percentile for RankIC with a floor of \(0.01\), a positive-day ratio of at least \(0.55\), and the 70th percentile for secondary metrics.
Candidates must also pass the elite rolling gate and benchmark-covered train--validation return checks requiring positive top-quantile benchmark-excess and long-short returns.
Warm-up retains the top \(15\%\) by elite score in place of static gating, with rolling and return checks still active.
Gate-passed candidates are truncated to a high-scoring prefix before archive selection.

\paragraph{Rolling gate requirements.}
OOS quality is summarized over all complete windows, with the local training-pass rate checked separately.
Table~\ref{tab:normal_elite_rolling_gates} lists the CSI300 thresholds.

\begin{table}[!htbp]
\centering
\small
\begin{tabular*}{\linewidth}{@{\extracolsep{\fill}}lcc@{}}
\toprule
Minimum requirement & Normal & Elite \\
\midrule
Total windows & 4 & -- \\
OOS windows & 3 & 5 \\
Training-pass rate & 0.25 & 0.55 \\
Mean aligned OOS RankIC & $-0.005$ & $0.01$ \\
OOS positive-window rate & 0.45 & 0.60 \\
Rolling RankICIR & $-0.05$ & $0.10$ \\
\bottomrule
\end{tabular*}
\caption{CSI300 rolling-OOS gate thresholds.
A dash indicates no separate total-window requirement.}
\label{tab:normal_elite_rolling_gates}
\end{table}

\paragraph{Parent and elite pools.}
\label{app:factor_pool_flow}
At initialization, the parent pool combines normal-selected seeds with raw and primitive baseline factors.
Each subsequent generation uses the previous elite archive together with a normal pool rebuilt from selected children and capped at 80 factors.
Pool selection retains a top-ranked head and fills the remaining slots through tournament selection.
Archive updates merge and deduplicate new elites within the configured capacity, reserving a quota for newly discovered factors.

\paragraph{Library score for portfolio evaluation.}
\label{app:library_score}
\label{app:cycle_final_candidate_diagnostics}
For final portfolio evaluation, elite factors are ranked by their train--validation predictive score, referred to as the library score:
\[
s_{\mathrm{library}}
=
0.70 \cdot \operatorname{Norm}(\mathrm{RankIC}^{\mathrm{aligned,winsor}}_{\mathrm{train+val}})
+
0.30 \cdot \operatorname{Norm}(\mathrm{IC}^{\mathrm{aligned,winsor}}_{\mathrm{train+val}}).
\]
The library score orders elite factors for correlation-constrained portfolio selection without a complexity discount or rolling-OOS term.
The selected subset and held-out reporting protocol are specified in Section~\ref{sec:experimental_settings}.

\paragraph{Interfaces to discovery and feedback.}
Discovery-stage redundancy screening follows Appendix~\ref{app:hypothesis_to_code}, with its correlation criterion defined in Appendix~\ref{app:redundancy_filtering}.
Final portfolio selection applies a separate correlation constraint to the archived elites without changing archive membership.

\section{Feedback and Memory Update}
\label{app:feedback_memory_update}

\paragraph{Factor-to-Archetype Attribution.}
Code-validated candidates are assigned mechanism labels before empirical selection.
Attribution uses the factor's code and explanatory fields to identify its mechanism family in the shared B/C taxonomy.
A second classification call selects a compatible \emph{Research Archetype} within that family.

\paragraph{Feedback Construction.}
Elite-selected factors supply GOOD records that connect their mechanisms to predictive evidence.
The LLM extracts a reusable mechanism principle and an avoid-copying constraint to guide new implementations of the underlying idea.
Normal-gate failures supply BAD records with the recorded rejection reasons.
These records explain the failed assumption and specify an avoidance rule or possible repair condition.

\paragraph{Memory Update and Reuse.}
Only elite-factor hypotheses are written to archetype-specific memory under their attributed \emph{Research Archetypes} to supplement fresh-seed planning.
Within the current cycle, short-lived GOOD/BAD summaries guide periodic novelty injection.
Its generation and screening procedure is described in Appendix~\ref{app:hypothesis_to_code}.
Cycle finalization consolidates positive and negative evidence under the cycle theme and explored archetypes into a persistent summary.
Later routing uses recent summaries to identify an unresolved information gap and formulate the next cycle theme (Appendix~\ref{app:memory_routing}).

\section{Metric Definitions}
\label{app:evaluation_details}

Evaluation and selection procedures are described in Appendix~\ref{app:factor_selection_details}; generation and novelty injection are described in Appendix~\ref{app:hypothesis_to_code}.

\subsection{Single-Factor Predictive Metrics}
\label{app:single_factor_metrics}

For each generated factor $f$, let $x_{i,t}^{(f)}$ denote the factor value of stock $i$ on day $t$, and let $r_{i,t+h}$ denote the future return at horizon $h$.
On each trading day $t$, \method{} computes cross-sectional predictive diagnostics over all valid stocks in the universe.
The default evaluation requires at least 20 valid stocks per day and nonconstant factor and return values.

The daily Information Coefficient (IC) is the Pearson correlation between factor values and future returns:
\begin{equation}
\mathrm{IC}_{t}^{(f)}
=
\mathrm{Corr}
\left(
x_{\cdot,t}^{(f)}, r_{\cdot,t+h}
\right).
\end{equation}

The daily Rank Information Coefficient (RankIC) is the Spearman rank correlation:
\begin{equation}
\mathrm{RankIC}_{t}^{(f)}
=
\mathrm{Corr}
\left(
\mathrm{rank}(x_{\cdot,t}^{(f)}),
\mathrm{rank}(r_{\cdot,t+h})
\right).
\end{equation}

For an evaluation split $s$, such as train, validation, or train--validation, the split-level IC and RankIC are the time-series averages:

\begin{equation}
    \mathcal{T}_{s}^{(f)}
=
\{t \in T_s : \mathrm{IC}_{t}^{(f)} \text{ and } \mathrm{RankIC}_{t}^{(f)} \text{ are valid}\}.
\end{equation}

\begin{equation}
\mathrm{IC}_{s}^{(f)}
=
\frac{1}{|\mathcal{T}_{s}^{(f)}|}
\sum_{t \in \mathcal{T}_{s}^{(f)}}
\mathrm{IC}_{t}^{(f)}.
\end{equation}

\begin{equation}
\mathrm{RankIC}_{s}^{(f)}
=
\frac{1}{|\mathcal{T}_{s}^{(f)}|}
\sum_{t \in \mathcal{T}_{s}^{(f)}}
\mathrm{RankIC}_{t}^{(f)}.
\end{equation}

The corresponding information ratios are defined as mean divided by standard deviation:
\begin{equation}
\mathrm{ICIR}_{s}^{(f)}
=
\frac{
\mathrm{mean}_{t \in \mathcal{T}_{s}^{(f)}}
\left(
\mathrm{IC}_{t}^{(f)}
\right)
}{
\mathrm{std}_{t \in \mathcal{T}_{s}^{(f)}}
\left(
\mathrm{IC}_{t}^{(f)}
\right)
}.
\end{equation}

\begin{equation}
\mathrm{RankICIR}_{s}^{(f)}
=
\frac{
\mathrm{mean}_{t \in \mathcal{T}_{s}^{(f)}}
\left(
\mathrm{RankIC}_{t}^{(f)}
\right)
}{
\mathrm{std}_{t \in \mathcal{T}_{s}^{(f)}}
\left(
\mathrm{RankIC}_{t}^{(f)}
\right)
}.
\end{equation}

Information ratios use sample standard deviations over valid dates and are defined for at least two dates with nonzero variance.
Winsorization clips factor values to the daily cross-sectional 0.5th and 99.5th percentiles and fills missing values with the cross-sectional median.
Selection uses direction-aligned winsorized metrics, with direction determined from raw training RankIC.

\subsection{Direction Alignment}
\label{app:direction_alignment}

A useful factor may be either positively or negatively associated with future returns.
To compare long and short signals under a single positive-is-better convention, \method{} fixes a factor direction using the raw training RankIC:
\begin{equation}
d_f =
\mathrm{sign}
\left(
\mathrm{RankIC}^{\mathrm{raw}}_{\mathrm{train}}(f)
\right),
\qquad
\mathrm{sign}(0)=1.
\end{equation}

For any metric $m \in \{\mathrm{IC}, \mathrm{RankIC}, \mathrm{ICIR}, \mathrm{RankICIR}\}$ and split $s$, the direction-aligned metric is:
\begin{equation}
\widetilde{m}_{s}^{(f)}
=
d_f
\cdot
m^{\mathrm{winsor}}_{s}(f).
\end{equation}

After alignment, larger values consistently indicate stronger predictive quality in the selected direction.

\subsection{Alpha Score}
\label{app:alpha_score}

The evaluator rescales aligned IC and RankIC before computing the alpha score:
\begin{equation}
\operatorname{Norm}(\widetilde{m}_{s}^{(f)})
=
15\widetilde{m}_{s}^{(f)},
\qquad m\in\{\mathrm{IC},\mathrm{RankIC}\}.
\end{equation}

The default alpha score gives RankIC a weight of 0.70 and IC a weight of 0.30:
\begin{equation}
S_{\alpha,s}^{(f)}
=
15\left(
0.70\widetilde{\mathrm{RankIC}}_{s}^{(f)}
+
0.30\widetilde{\mathrm{IC}}_{s}^{(f)}
\right).
\end{equation}

ICIR and RankICIR support separate stability checks, with temporal robustness assessed by rolling-OOS gates.

\subsection{Complexity Discount}
\label{app:complexity_discount}

\method{} scores the generated function's AST to control structural complexity.
The analyzer groups structural features into the following component scores:
\begin{equation}
\begin{aligned}
C_{\mathrm{expr}} &: \text{expression and intermediate-variable complexity},\\
C_{\mathrm{flow}} &: \text{branching and loop complexity},\\
C_{\mathrm{ts}} &: \text{rolling, expanding, EWM, and nested time-series structure},\\
C_{\mathrm{param}} &: \text{numeric-threshold and window-parameter risk},\\
C_{\mathrm{stat}} &: \text{statistical-operator risk},\\
C_{\mathrm{chain}} &: \text{long transformation-chain complexity}.
\end{aligned}
\end{equation}
The component scores are aggregated into an implementation-burden score and an overfit-risk score:
\begin{equation}
\begin{aligned}
C_{\mathrm{impl}}^{(f)}
&=
C_{\mathrm{expr}}^{(f)}
+
C_{\mathrm{flow}}^{(f)}
+
0.25 C_{\mathrm{ts}}^{(f)}
+
0.45 C_{\mathrm{chain}}^{(f)}
+
C_{\mathrm{impl,stat}}^{(f)},\\
C_{\mathrm{risk}}^{(f)}
&=
C_{\mathrm{param}}^{(f)}
+
C_{\mathrm{stat}}^{(f)}
+
0.75 C_{\mathrm{ts}}^{(f)}
+
0.65 C_{\mathrm{chain}}^{(f)}
+
C_{\mathrm{risk,flow}}^{(f)}.
\end{aligned}
\end{equation}
Here $C_{\mathrm{impl,stat}}^{(f)}$ contains the implementation burden of medium-risk statistical operators, while $C_{\mathrm{risk,flow}}^{(f)}$ contains additional overfit-risk penalties for explicit branches and loops.
The raw total complexity reported in factor examples is then:
\begin{equation}
C^{(f)}
=
0.40 C_{\mathrm{impl}}^{(f)}
+
0.60 C_{\mathrm{risk}}^{(f)}.
\end{equation}

The score is converted into a multiplicative piecewise discount:
\begin{equation}
D_{\mathrm{raw}}^{(f)}
=
\begin{cases}
0.9 + 0.1(10.8-C^{(f)})/10.8, & C^{(f)} \le 10.8,\\
0.9 - 0.15(C^{(f)}-10.8)/(15.0-10.8), & 10.8 < C^{(f)} \le 15.0,\\
0.75\exp[-0.25(C^{(f)}-15.0)], & C^{(f)} > 15.0.
\end{cases}
\end{equation}
The final complexity discount is floored at $0.70$:
\begin{equation}
D_{\mathrm{cmp}}^{(f)}
=
\max
\left(
0.70,
D_{\mathrm{raw}}^{(f)}
\right).
\end{equation}

In the main setting, the discount multiplies the normal-selection score as defined in Appendix~\ref{app:factor_selection_details}.
Elite selection and the final library score use predictive quality without an additional complexity discount (Appendix~\ref{app:factor_selection_details}).

\subsection{Rolling Out-of-Sample Gate}
\label{app:rolling_oos_gate}

Static split performance can overstate factor quality.
\method{} therefore evaluates each factor with rolling out-of-sample windows.
Each rolling window $j$ contains a training segment $T_j^{\mathrm{train}}$ followed by an out-of-sample segment $T_j^{\mathrm{oos}}$.
The window uses a five-year training segment, a six-month OOS segment, and a three-month step.
Screening windows are constructed within the train--validation period, excluding held-out test dates.
Only complete windows with nonempty training and OOS segments are included.

For each rolling training segment, the evaluator computes local aligned RankIC, aligned RankICIR, and positive-day ratio.
Let $\rho_j^{+}$ denote the fraction of positive values among finite aligned RankIC observations whose absolute magnitude exceeds $\epsilon=10^{-8}$.
The rolling training segment passes the local stability gate if:
\begin{equation}
\overline{
\widetilde{\mathrm{RankIC}}
}_{j,\mathrm{train}}
\ge
\tau_{\mathrm{rankic}}^{\mathrm{train}}.
\end{equation}

\begin{equation}
\widetilde{\mathrm{RankICIR}}_{j,\mathrm{train}}
\ge
\tau_{\mathrm{rankicir}}^{\mathrm{train}}.
\end{equation}

\begin{equation}
\rho_{j,\mathrm{train}}^{+}
\ge
\tau_{\rho}^{\mathrm{train}}.
\end{equation}
The default local thresholds are $0.005$ for mean aligned RankIC, $0.05$ for RankICIR, and $0.55$ for the positive-day ratio.

Let $\mathcal{J}_{\mathrm{all}}$ denote all complete rolling windows with non-empty OOS observations, and let $\mathcal{J}_{\mathrm{pass}}$ denote the subset whose preceding training segments pass the local stability gate.
The train-pass rate is
\begin{equation}
\rho_{\mathrm{train}}^{\mathrm{pass}}
=
\frac{
|\mathcal{J}_{\mathrm{pass}}|
}{
|\mathcal{J}_{\mathrm{all}}|
}.
\end{equation}
The factor-level hard gate uses OOS summaries over $\mathcal{J}_{\mathrm{all}}$ and checks the training-pass rate separately.
OOS summaries over $\mathcal{J}_{\mathrm{pass}}$ are retained as additional diagnostics.
The mean aligned OOS RankIC is
\begin{equation}
\overline{
\widetilde{\mathrm{RankIC}}
}_{\mathrm{oos}}
=
\frac{1}{
|\mathcal{J}_{\mathrm{all}}|
}
\sum_{j \in \mathcal{J}_{\mathrm{all}}}
\overline{
\widetilde{\mathrm{RankIC}}
}_{j,\mathrm{oos}}.
\end{equation}

Let $\mathcal{J}_{\mathrm{nz}}$ contain the windows in $\mathcal{J}_{\mathrm{all}}$ whose absolute mean aligned OOS RankIC exceeds $\epsilon$.
The positive-window rate excludes near-zero window means:
\begin{equation}
\rho_{\mathrm{oos}}^{+}
=
\frac{1}{
|\mathcal{J}_{\mathrm{nz}}|
}
\sum_{j \in \mathcal{J}_{\mathrm{nz}}}
\mathbf{1}
\left[
\overline{
\widetilde{\mathrm{RankIC}}
}_{j,\mathrm{oos}}
>
0
\right].
\end{equation}
The rate is zero when $\mathcal{J}_{\mathrm{nz}}$ is empty.

Rolling RankICIR summarizes variation across all window means:
\begin{equation}
\widetilde{\mathrm{RankICIR}}_{\mathrm{oos}}
=
\frac{
\mathrm{mean}_{j \in \mathcal{J}_{\mathrm{all}}}
\left(
\overline{
\widetilde{\mathrm{RankIC}}
}_{j,\mathrm{oos}}
\right)
}{
\mathrm{std}_{j \in \mathcal{J}_{\mathrm{all}}}
\left(
\overline{
\widetilde{\mathrm{RankIC}}
}_{j,\mathrm{oos}}
\right)
+ \epsilon
}.
\end{equation}
The standard deviation uses the sample convention, and the ratio is clipped to $[-10,10]$.
It is set to zero when OOS coverage falls below the normal gate's minimum window count.

In the main setting, rolling diagnostics determine eligibility through the normal and elite hard gates in Appendix~\ref{app:factor_selection_details}.
These factor-screening windows are separate from the rolling portfolio-model fitting protocol in Section~\ref{sec:experimental_settings}.

\subsection{Factor-Correlation Criterion}
\label{app:redundancy_filtering}

Generation-level redundancy screening uses training-period factor values, winsorized within each daily cross-section.
For factors $f_a$ and $f_b$, let $\mathcal{T}_{a,b}$ contain dates with valid cross-sectional Pearson correlations.
Their correlation is the signed daily mean:
\begin{equation}
\rho(f_a,f_b)
=
\frac{1}{|\mathcal{T}_{a,b}|}
\sum_{t\in\mathcal{T}_{a,b}}
\operatorname{Corr}_{i}
\left(x_{i,t}^{(f_a)},x_{i,t}^{(f_b)}\right).
\end{equation}
For a comparison pool $\mathcal{S}$, a candidate passes when:
\begin{equation}
\max_{f' \in \mathcal{S}}
\left|
\rho(f,f')
\right|
<
\tau_{\mathrm{corr}}.
\end{equation}

In the default setting, the generation-level correlation threshold is $\tau_{\mathrm{corr}}=0.95$.
The absolute value is taken after averaging the daily correlations.
For a nonempty comparison pool, candidates with fewer than 120 valid comparison days or less than 30\% date coverage are rejected conservatively.
Its use during discovery is described in Appendix~\ref{app:hypothesis_to_code}; portfolio-factor selection follows Section~\ref{sec:experimental_settings}.

\subsection{Portfolio and Backtest Metrics}
\label{app:portfolio_metrics}

We compute annualized return (AR), annualized excess return (AER), and information ratio (IR) using Qlib's TopK-Dropout backtest and risk-analysis routine.
Let $r_t^{\mathrm{port}}$ denote the daily portfolio return, $r_t^{\mathrm{bench}}$ the benchmark return, and $c_t$ the transaction cost on day $t$.
Following the Qlib backtest convention, the cost-adjusted daily excess return is
\begin{equation}
r_t^{\mathrm{excess}}
=
r_t^{\mathrm{port}}
-
r_t^{\mathrm{bench}}
-
c_t .
\end{equation}

For daily backtests, Qlib's day-frequency annualization scaler $N$ is used, which is 238 under the day-frequency default used by \texttt{risk\_analysis}.
Under the product-mode risk-analysis routine used in our implementation, the compounded curves are
\begin{equation}
\Pi_T^{\mathrm{port}}
=
\prod_{t=1}^{T}
\left(1+r_t^{\mathrm{port}}\right),
\qquad
\Pi_T^{\mathrm{excess}}
=
\prod_{t=1}^{T}
\left(1+r_t^{\mathrm{excess}}\right).
\end{equation}
AR and AER are then reported as the corresponding compounded annualized returns:
\begin{equation}
\mathrm{AR}
=
\left(
\Pi_T^{\mathrm{port}}
\right)^{N/T}
-
1,
\qquad
\mathrm{AER}
=
\left(
\Pi_T^{\mathrm{excess}}
\right)^{N/T}
-
1.
\end{equation}

AR uses portfolio returns before transaction costs; AER uses cost-adjusted excess returns.
In Qlib's product mode, IR uses the geometric mean daily excess return and sample standard deviation of log excess returns:
\begin{equation}
\mathrm{IR}
=
\frac{
\left(\Pi_T^{\mathrm{excess}}\right)^{1/T}-1
}{
\mathrm{std}_{t}\left[\log\left(1+r_t^{\mathrm{excess}}\right)\right]
}
\sqrt{N}.
\end{equation}

\section{Experimental Protocol Details}
\label{app:experimental_protocol}

\begin{figure}[t]
\centering
\includegraphics[width=0.95\textwidth]{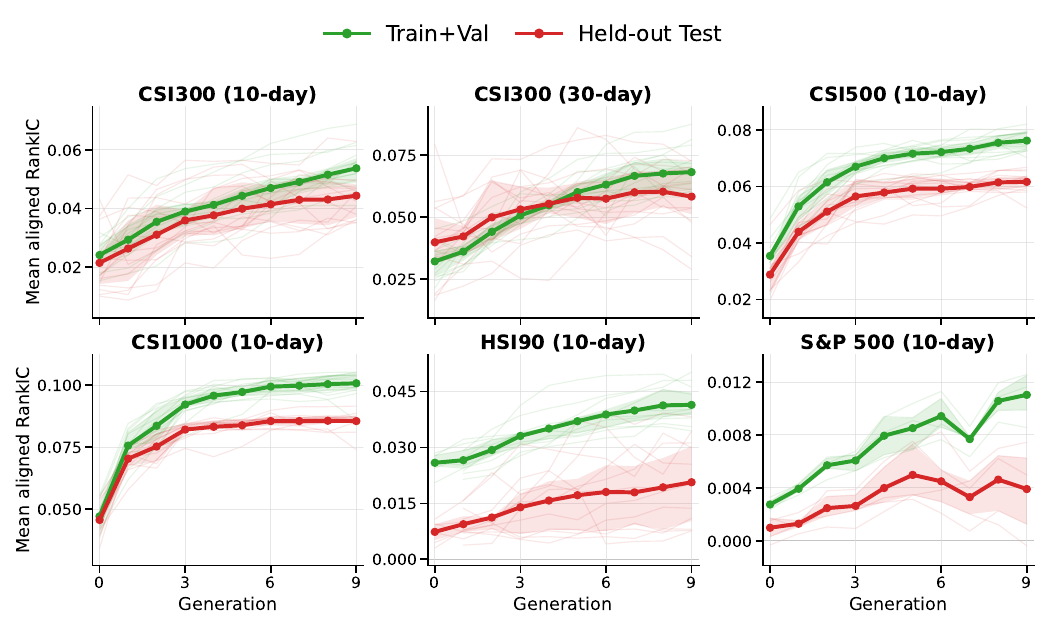}
\vspace{-0.2cm}
\caption{Complete generation-level direction-aligned RankIC across all six settings.
Visual encoding follows Figure~\ref{fig:cross_market_generation_rankic}; vertical scales are panel-specific.}
\label{fig:cross_market_generation_rankic_full}
\end{figure}

\begin{table*}[!htbp]
\centering
\scriptsize
\setlength{\tabcolsep}{3.2pt}
\renewcommand{\arraystretch}{1.02}
\resizebox{\textwidth}{!}{
\begin{tabular}{@{}llrrrrrrr@{}}
\toprule
\textbf{Type} & \textbf{Model} & \textbf{IC} & \textbf{RankIC} & \textbf{ICIR} & \textbf{RankICIR} & \textbf{AR} & \textbf{AER} & \textbf{IR} \\
\midrule
\multirow{6}{*}{Machine Learning}
& AdaBoost & $0.0161\pm0.0007$ & $0.0191\pm0.0011$ & $0.2051\pm0.0080$ & $0.2054\pm0.0111$ & $0.0536\pm0.0079$ & $0.0457\pm0.0077$ & $0.6714\pm0.1124$ \\
& CatBoost & $0.0086\pm0.0009$ & $0.0111\pm0.0006$ & $0.0963\pm0.0102$ & $0.1393\pm0.0080$ & $0.0554\pm0.0094$ & $0.0468\pm0.0094$ & $0.7391\pm0.1580$ \\
& LightGBM & $0.0213\pm0.0011$ & $0.0291\pm0.0007$ & $0.2180\pm0.0109$ & $0.2454\pm0.0056$ & $0.0683\pm0.0168$ & $0.0612\pm0.0165$ & $0.9003\pm0.2259$ \\
& Random Forest & $0.0117\pm0.0002$ & $0.0141\pm0.0003$ & $0.1391\pm0.0022$ & $0.1680\pm0.0032$ & $0.0553\pm0.0032$ & $0.0467\pm0.0032$ & $0.7010\pm0.0414$ \\
& Ridge & $0.0118\pm0.0000$ & $0.0157\pm0.0000$ & $0.1249\pm0.0000$ & $0.1318\pm0.0000$ & $0.0004\pm0.0000$ & $-0.0070\pm0.0000$ & $-0.0978\pm0.0000$ \\
& XGBoost & $0.0161\pm0.0001$ & $0.0162\pm0.0001$ & $0.1942\pm0.0018$ & $0.1894\pm0.0013$ & $0.0564\pm0.0035$ & $0.0475\pm0.0035$ & $0.7167\pm0.0573$ \\
\midrule
\multirow{5}{*}{Deep Learning}
& CNN & $0.0152\pm0.0032$ & $0.0226\pm0.0022$ & $0.1380\pm0.0278$ & $0.1995\pm0.0199$ & $0.0481\pm0.0119$ & $0.0381\pm0.0120$ & $0.4886\pm0.1431$ \\
& GRU & $0.0107\pm0.0051$ & $0.0225\pm0.0042$ & $0.1256\pm0.0605$ & $0.1983\pm0.0325$ & $0.0544\pm0.0068$ & $0.0442\pm0.0061$ & $0.5647\pm0.0728$ \\
& LSTM & $0.0112\pm0.0033$ & $0.0240\pm0.0030$ & $0.1308\pm0.0380$ & $0.2115\pm0.0249$ & $0.0517\pm0.0066$ & $0.0414\pm0.0072$ & $0.5261\pm0.0956$ \\
& MLP & $0.0156\pm0.0068$ & $0.0237\pm0.0078$ & $0.1415\pm0.0596$ & $0.2043\pm0.0587$ & $0.0419\pm0.0226$ & $0.0331\pm0.0220$ & $0.4305\pm0.2849$ \\
& Transformer & $0.0079\pm0.0099$ & $0.0150\pm0.0083$ & $0.0749\pm0.0924$ & $0.1365\pm0.0721$ & $0.0473\pm0.0215$ & $0.0383\pm0.0214$ & $0.5459\pm0.3059$ \\
\midrule
\multirow{5}{*}{Alpha Methods}
& Alpha360 & $0.0129\pm0.0000$ & $0.0099\pm0.0000$ & $0.0855\pm0.0000$ & $0.0636\pm0.0000$ & $0.0317\pm0.0000$ & $0.0299\pm0.0000$ & $0.2793\pm0.0000$ \\
& AlphaAgent & $0.0198\pm0.0034$ & $0.0233\pm0.0037$ & $0.1389\pm0.0262$ & $0.1616\pm0.0264$ & $0.0534\pm0.0063$ & $0.0459\pm0.0062$ & $0.5222\pm0.0592$ \\
& AutoAgent & $0.0240\pm0.0035$ & $0.0301\pm0.0054$ & $0.1514\pm0.0187$ & $0.1843\pm0.0313$ & $0.0827\pm0.0116$ & $0.0756\pm0.0111$ & $0.6961\pm0.1032$ \\
& CogAlpha & $0.0297\pm0.0075$ & $0.0373\pm0.0080$ & $0.1796\pm0.0414$ & $0.2141\pm0.0419$ & $0.0649\pm0.0356$ & $0.0562\pm0.0357$ & $0.6088\pm0.4550$ \\
& R\&D-Agent(Q) & $0.0261\pm0.0037$ & $0.0334\pm0.0045$ & $0.1801\pm0.0332$ & $0.2361\pm0.0289$ & $0.0756\pm0.0205$ & $0.0669\pm0.0209$ & $0.7790\pm0.2437$ \\
\midrule
\multirow{6}{*}{Single LLM}
& DeepSeek-v3.2 & $0.0220\pm0.0132$ & $0.0225\pm0.0175$ & $0.1674\pm0.0901$ & $0.1642\pm0.1195$ & $0.0684\pm0.0355$ & $0.0599\pm0.0339$ & $0.7197\pm0.4235$ \\
& Gemma-4-31B & $0.0328\pm0.0077$ & $0.0367\pm0.0098$ & $0.2440\pm0.0652$ & $0.2586\pm0.0731$ & $0.0826\pm0.0257$ & $0.0736\pm0.0260$ & $0.8587\pm0.3024$ \\
& GLM-5.2 & $0.0132\pm0.0081$ & $0.0136\pm0.0081$ & $0.1056\pm0.0652$ & $0.1039\pm0.0622$ & $0.0508\pm0.0180$ & $0.0441\pm0.0171$ & $0.5262\pm0.2238$ \\
& GPT-OSS-120B & $0.0243\pm0.0115$ & $0.0303\pm0.0141$ & $0.1583\pm0.0737$ & $0.2004\pm0.0924$ & $0.0600\pm0.0369$ & $0.0518\pm0.0370$ & $0.5476\pm0.3911$ \\
& Kimi-K2.7-Code & $0.0240\pm0.0029$ & $0.0257\pm0.0040$ & $0.1891\pm0.0172$ & $0.1918\pm0.0199$ & $0.0636\pm0.0097$ & $0.0565\pm0.0097$ & $0.6336\pm0.1077$ \\
& MiniMax-M2.5 & $0.0270\pm0.0045$ & $0.0278\pm0.0062$ & $0.2069\pm0.0311$ & $0.2002\pm0.0331$ & $0.0741\pm0.0260$ & $0.0663\pm0.0240$ & $0.7689\pm0.2861$ \\
\midrule
Ours & \method{} & $\mathbf{0.0448}\pm0.0118$ & $\mathbf{0.0590}\pm0.0113$ & $\mathbf{0.2602}\pm0.0515$ & $\mathbf{0.3379}\pm0.0303$ & $\mathbf{0.1379}\pm0.0345$ & $\mathbf{0.1291}\pm0.0328$ & $\mathbf{1.3472}\pm0.3105$ \\
\bottomrule
\end{tabular}
}
\vspace{-0.2cm}
\caption{Complete CSI300 10-day rolling comparison.
Values are mean $\pm$ SD across five outcomes per method.
The best mean for each metric is bolded.}
\label{tab:complete_baseline_rolling}
\vspace{-0.2cm}
\end{table*}

\paragraph{Data source and universe.}
The CSI300, CSI500, and CSI1000 adjusted daily OHLCV datasets are obtained from Qlib~\citep{yang2020qlib}.
SP500 and HSI90 data are obtained from Yahoo Finance~\citep{Aroussi2024} and organized in the same Qlib-compatible format.
We evaluate CSI300 at 10 and 30 days and the remaining indices at 10 days, with baseline comparisons and component studies focused on CSI300 at 10 days.
Financial reports used by RMA are collected primarily from FXBaogao (\url{https://www.fxbaogao.com/}).

\paragraph{Calendar splits.}
We use 2011-01-01--2020-12-31 for training, 2021-01-01--2021-12-31 for validation, and 2022-01-01--2025-12-31 for testing.
During factor discovery, training and validation provide selection evidence; test metrics are reserved for reporting.

\paragraph{Target construction and preprocessing.}
All methods share a \texttt{(date,ticker)} OHLCV panel with targets defined in Section~\ref{sec:experimental_settings}.
Duplicates retain their last observation, and rows with missing identifiers, targets, or OHLCV values are dropped after non-finite values are marked missing.
A 540-calendar-day buffer before training supplies historical observations for factor computation.
We require positive prices, volume of at least 1, and a close-price--volume product of at least 1,000 before dividing volume by $10^6$.
Each ticker must provide at least $\min\{300,\max\{120,\lceil0.20D\rceil\}\}$ valid observations between training start and evaluation end, where $D$ counts available trading dates in that period.

\paragraph{Model fitting.}
All methods in the main comparison use expanding-history models refitted every 128 trading days to predict successive non-overlapping windows.
The static comparison in Appendix~\ref{app:baseline_static_results} fits each model once on pre-test history and keeps it fixed throughout the test period.
Both protocols apply an $h+1$-trading-day embargo before each prediction window, excluding those dates from fitting (11 days for the 10-day target).
Rolling fits may include earlier test-period observations once their forward-return labels are available.
Generated-factor groups are fixed before testing and fitted with Ridge regression~\citep{hoerl1970ridge} using $\alpha=10.0$.

\paragraph{Portfolio backtesting.}
We use Qlib~\citep{yang2020qlib} with an initial capital of CNY 100 million for the CSI300 portfolio comparison.
The top-$50$/drop-$5$ strategy targets 50 stocks ranked by predicted return, with up to 5 daily replacements executed at the opening price.
Fixed buy and sell cost rates of 0.026\% and 0.076\% combine a 0.025\% brokerage commission and a 0.001\% transfer fee on each side, plus a 0.05\% stamp duty on sales.
Qlib applies a minimum transaction cost of CNY 5 per order to approximate minimum brokerage charges.
AR uses gross portfolio returns, and excess-return metrics use the CSI300 benchmark after transaction costs (Appendix~\ref{app:portfolio_metrics}).

\paragraph{Main runtime configuration.}
The default LLM is \texttt{gpt-oss-120b}~\citep{openai2025gptoss}, with LLM-based hypothesis-subset planning enabled.
Each research cycle targets 64 initial factors and retains up to 80 normal-selected parents across 10 generations.
The novelty-injection schedule and validation settings are specified in Appendix~\ref{app:hypothesis_to_code}.
Normal and elite screening thresholds and pool-selection rules are given in Appendix~\ref{app:factor_selection_details}.

\paragraph{Cross-market generation analysis.}
Figure~\ref{fig:cross_market_generation_rankic} presents four representative settings, and Figure~\ref{fig:cross_market_generation_rankic_full} includes all six settings under the ten-generation research schedule.
For each generation, we average RankIC over the top 8 elite factors within each cycle, then average the cycle-level means across 15 cycles.
Approximately 100 elite factors contribute to each generation-level estimate.
These substantial relative gains across all evaluated markets support the cross-market generalization of \method{}'s factor-discovery capability.

\begin{table*}[t]
\centering
\scriptsize
\setlength{\tabcolsep}{12pt}
\renewcommand{\arraystretch}{1.02}
% \resizebox{\textwidth}{!}{
\begin{tabular}{@{}llrrrrrl@{}}
\toprule
\textbf{Type} & \textbf{Model} & \textbf{01} & \textbf{02} & \textbf{03} & \textbf{04} & \textbf{05} & \textbf{Mean AER} \\
\midrule
\multirow{6}{*}{Machine Learning}
& AdaBoost & 0.0491 & 0.0349 & 0.0405 & 0.0516 & 0.0525 & 0.0457 $\pm$ 0.0077 \\
& CatBoost & 0.0543 & 0.0398 & 0.0349 & 0.0569 & 0.0479 & 0.0468 $\pm$ 0.0094 \\
& LightGBM & 0.0522 & 0.0900 & 0.0529 & 0.0509 & 0.0598 & 0.0612 $\pm$ 0.0165 \\
& Random Forest & 0.0460 & 0.0508 & 0.0420 & 0.0483 & 0.0467 & 0.0467 $\pm$ 0.0032 \\
& Ridge & -0.0070 & -0.0070 & -0.0070 & -0.0070 & -0.0070 & -0.0070 $\pm$ 0.0000 \\
& XGBoost & 0.0513 & 0.0468 & 0.0440 & 0.0509 & 0.0445 & 0.0475 $\pm$ 0.0035 \\
\midrule
\multirow{5}{*}{Deep Learning}
& CNN & 0.0464 & 0.0402 & 0.0196 & 0.0503 & 0.0342 & 0.0381 $\pm$ 0.0120 \\
& GRU & 0.0439 & 0.0438 & 0.0538 & 0.0422 & 0.0370 & 0.0442 $\pm$ 0.0061 \\
& LSTM & 0.0321 & 0.0416 & 0.0524 & 0.0402 & 0.0406 & 0.0414 $\pm$ 0.0072 \\
& MLP & 0.0463 & -0.0042 & 0.0516 & 0.0335 & 0.0383 & 0.0331 $\pm$ 0.0220 \\
& Transformer & 0.0008 & 0.0414 & 0.0500 & 0.0531 & 0.0463 & 0.0383 $\pm$ 0.0214 \\
\midrule
\multirow{5}{*}{Alpha Methods}
& Alpha360 & 0.0299 & 0.0299 & 0.0299 & 0.0299 & 0.0299 & 0.0299 $\pm$ 0.0000 \\
& AlphaAgent & 0.0430 & 0.0499 & 0.0370 & 0.0464 & 0.0529 & 0.0459 $\pm$ 0.0062 \\
& AutoAgent & 0.0811 & 0.0578 & 0.0875 & 0.0752 & 0.0764 & 0.0756 $\pm$ 0.0111 \\
& CogAlpha & 0.0060 & 0.1021 & 0.0489 & 0.0752 & 0.0490 & 0.0562 $\pm$ 0.0357 \\
& R\&D-Agent(Q) & 0.0460 & 0.0722 & 0.0477 & 0.0715 & 0.0969 & 0.0669 $\pm$ 0.0209 \\
\midrule
\multirow{6}{*}{Single LLM}
& DeepSeek-v3.2 & 0.0930 & 0.0719 & 0.0866 & 0.0177 & 0.0302 & 0.0599 $\pm$ 0.0339 \\
& Gemma-4-31B & 0.0797 & 0.0777 & 0.0525 & 0.1119 & 0.0463 & 0.0736 $\pm$ 0.0260 \\
& GLM-5.2 & 0.0341 & 0.0623 & 0.0463 & 0.0205 & 0.0575 & 0.0441 $\pm$ 0.0171 \\
& GPT-OSS-120B & 0.0768 & 0.0003 & 0.0809 & 0.0244 & 0.0765 & 0.0518 $\pm$ 0.0370 \\
& Kimi-K2.7-Code & 0.0518 & 0.0548 & 0.0588 & 0.0714 & 0.0455 & 0.0565 $\pm$ 0.0097 \\
& MiniMax-M2.5 & 0.0904 & 0.0689 & 0.0282 & 0.0616 & 0.0822 & 0.0663 $\pm$ 0.0240 \\
\midrule
Ours & \method{} & 0.1599 & 0.1096 & 0.0906 & 0.1662 & 0.1191 & \textbf{0.1291} $\pm$ 0.0328 \\
\bottomrule
\end{tabular}
% }
\vspace{-0.2cm}
\caption{Individual rolling-AER outcomes for all evaluated methods.
Columns 01--05 report the five seed- or group-level values; the final column reports their mean $\pm$ sample standard deviation.
The highest mean AER is bolded.}
\label{tab:baseline_group_aer}
\vspace{-0.2cm}
\end{table*}

\section{Complete Baseline Results}
\label{app:baseline_results}

\paragraph{Aggregation rule.}
We report means and sample standard deviations of all seven metrics across five outcomes per method.
The outcomes are random seeds for ML/DL models and Alpha360, or non-overlapping 40-factor groups for generated-factor methods, including \method{}.
Table~\ref{tab:complete_baseline_rolling} provides the complete rolling comparison, and Table~\ref{tab:baseline_group_aer} lists the individual AER values.

\paragraph{Factor-group construction.}
Each generated-factor method provides 1,500 valid candidates, with the \method{} pool consisting of elite factors.
Before test evaluation, train--validation library-score ranking (Appendix~\ref{app:library_score}) forms five non-overlapping 40-factor groups with pairwise absolute correlation below $0.60$.
Each group is built by scanning candidates in descending score order and excluding factors selected for earlier groups.
Ridge fitting follows the rolling and static protocols in Appendix~\ref{app:experimental_protocol}.
Alpha360 is deterministic in this setting, so its five seed-level outcomes are identical.

\begin{table*}[t]
\centering
\scriptsize
\setlength{\tabcolsep}{3.2pt}
\renewcommand{\arraystretch}{1.02}
\resizebox{\textwidth}{!}{
\begin{tabular}{@{}llrrrrrrr@{}}
\toprule
\textbf{Type} & \textbf{Model} & \textbf{IC} & \textbf{RankIC} & \textbf{ICIR} & \textbf{RankICIR} & \textbf{AR} & \textbf{AER} & \textbf{IR} \\
\midrule
\multirow{6}{*}{Machine Learning}
& AdaBoost & $0.0142\pm0.0005$ & $0.0153\pm0.0004$ & $0.1778\pm0.0066$ & $0.1774\pm0.0037$ & $0.0436\pm0.0099$ & $0.0350\pm0.0100$ & $0.5767\pm0.1555$ \\
& CatBoost & $0.0008\pm0.0005$ & $0.0034\pm0.0006$ & $0.0089\pm0.0054$ & $0.0445\pm0.0075$ & $0.0340\pm0.0159$ & $0.0260\pm0.0157$ & $0.4569\pm0.2668$ \\
& LightGBM & $0.0167\pm0.0005$ & $0.0320\pm0.0015$ & $0.1853\pm0.0072$ & $0.2673\pm0.0035$ & $0.0666\pm0.0101$ & $0.0607\pm0.0100$ & $0.8825\pm0.1341$ \\
& Random Forest & $0.0060\pm0.0003$ & $0.0054\pm0.0001$ & $0.0777\pm0.0039$ & $0.0790\pm0.0020$ & $0.0259\pm0.0094$ & $0.0186\pm0.0095$ & $0.3226\pm0.1585$ \\
& Ridge & $0.0013\pm0.0000$ & $0.0053\pm0.0000$ & $0.0121\pm0.0000$ & $0.0396\pm0.0000$ & $0.0041\pm0.0000$ & $-0.0039\pm0.0000$ & $-0.0600\pm0.0000$ \\
& XGBoost & $0.0052\pm0.0001$ & $0.0060\pm0.0003$ & $0.0661\pm0.0019$ & $0.0861\pm0.0037$ & $0.0319\pm0.0022$ & $0.0237\pm0.0021$ & $0.4338\pm0.0380$ \\
\midrule
\multirow{5}{*}{Deep Learning}
& CNN & $0.0137\pm0.0077$ & $0.0202\pm0.0099$ & $0.1252\pm0.0812$ & $0.1759\pm0.0804$ & $0.0566\pm0.0158$ & $0.0464\pm0.0166$ & $0.6055\pm0.1855$ \\
& GRU & $0.0132\pm0.0013$ & $0.0268\pm0.0024$ & $0.1569\pm0.0132$ & $0.2242\pm0.0075$ & $0.0538\pm0.0118$ & $0.0434\pm0.0127$ & $0.5476\pm0.1549$ \\
& LSTM & $0.0116\pm0.0004$ & $0.0242\pm0.0006$ & $0.1406\pm0.0045$ & $0.2146\pm0.0027$ & $0.0475\pm0.0031$ & $0.0365\pm0.0031$ & $0.4653\pm0.0387$ \\
& MLP & $0.0142\pm0.0040$ & $0.0214\pm0.0058$ & $0.1350\pm0.0279$ & $0.1982\pm0.0476$ & $0.0468\pm0.0054$ & $0.0369\pm0.0055$ & $0.5449\pm0.1018$ \\
& Transformer & $-0.0010\pm0.0067$ & $0.0110\pm0.0057$ & $-0.0077\pm0.0554$ & $0.0974\pm0.0516$ & $0.0292\pm0.0127$ & $0.0191\pm0.0112$ & $0.3046\pm0.1739$ \\
\midrule
\multirow{5}{*}{Alpha Methods}
& Alpha360 & $0.0129\pm0.0000$ & $0.0070\pm0.0000$ & $0.0837\pm0.0000$ & $0.0433\pm0.0000$ & $0.0221\pm0.0000$ & $0.0226\pm0.0000$ & $0.1935\pm0.0000$ \\
& AlphaAgent & $0.0189\pm0.0035$ & $0.0237\pm0.0034$ & $0.1277\pm0.0241$ & $0.1585\pm0.0215$ & $0.0560\pm0.0070$ & $0.0484\pm0.0074$ & $0.5459\pm0.0633$ \\
& AutoAgent & $0.0167\pm0.0034$ & $0.0201\pm0.0053$ & $0.1013\pm0.0193$ & $0.1165\pm0.0310$ & $0.0659\pm0.0130$ & $0.0591\pm0.0124$ & $0.5406\pm0.1140$ \\
& CogAlpha & $0.0315\pm0.0062$ & $0.0406\pm0.0060$ & $0.1916\pm0.0347$ & $0.2309\pm0.0329$ & $0.0380\pm0.0241$ & $0.0256\pm0.0228$ & $0.3049\pm0.2728$ \\
& R\&D-Agent(Q) & $0.0255\pm0.0029$ & $0.0334\pm0.0035$ & $0.1713\pm0.0252$ & $0.2280\pm0.0186$ & $0.0769\pm0.0201$ & $0.0683\pm0.0210$ & $0.7943\pm0.2423$ \\
\midrule
\multirow{6}{*}{Single LLM}
& DeepSeek-v3.2 & $0.0215\pm0.0122$ & $0.0218\pm0.0162$ & $0.1642\pm0.0856$ & $0.1602\pm0.1140$ & $0.0577\pm0.0317$ & $0.0498\pm0.0304$ & $0.5999\pm0.3811$ \\
& Gemma-4-31B & $0.0295\pm0.0085$ & $0.0328\pm0.0094$ & $0.2268\pm0.0765$ & $0.2379\pm0.0729$ & $0.0864\pm0.0244$ & $0.0783\pm0.0251$ & $0.9110\pm0.2907$ \\
& GLM-5.2 & $0.0129\pm0.0074$ & $0.0130\pm0.0073$ & $0.1035\pm0.0602$ & $0.0999\pm0.0574$ & $0.0474\pm0.0200$ & $0.0413\pm0.0196$ & $0.4722\pm0.2283$ \\
& GPT-OSS-120B & $0.0214\pm0.0118$ & $0.0267\pm0.0150$ & $0.1398\pm0.0773$ & $0.1755\pm0.0991$ & $0.0445\pm0.0311$ & $0.0362\pm0.0314$ & $0.3862\pm0.3412$ \\
& Kimi-K2.7-Code & $0.0216\pm0.0026$ & $0.0226\pm0.0036$ & $0.1731\pm0.0135$ & $0.1712\pm0.0165$ & $0.0603\pm0.0156$ & $0.0536\pm0.0159$ & $0.5996\pm0.1749$ \\
& MiniMax-M2.5 & $0.0261\pm0.0052$ & $0.0276\pm0.0070$ & $0.2007\pm0.0370$ & $0.1988\pm0.0409$ & $0.0642\pm0.0307$ & $0.0563\pm0.0290$ & $0.6709\pm0.3819$ \\
\midrule
Ours & \method{} & $\mathbf{0.0458}\pm0.0119$ & $\mathbf{0.0584}\pm0.0119$ & $\mathbf{0.2794}\pm0.0620$ & $\mathbf{0.3466}\pm0.0364$ & $\mathbf{0.1233}\pm0.0420$ & $\mathbf{0.1146}\pm0.0404$ & $\mathbf{1.2097}\pm0.4014$ \\
\bottomrule
\end{tabular}
}
\caption{Complete CSI300 10-day static-fit comparison.
Values are mean $\pm$ SD across five outcomes per method.
The best mean for each metric is bolded.}
\label{tab:complete_baseline_static}
\end{table*}

\begin{figure}[!htbp]
\centering
\includegraphics[width=0.95\textwidth]{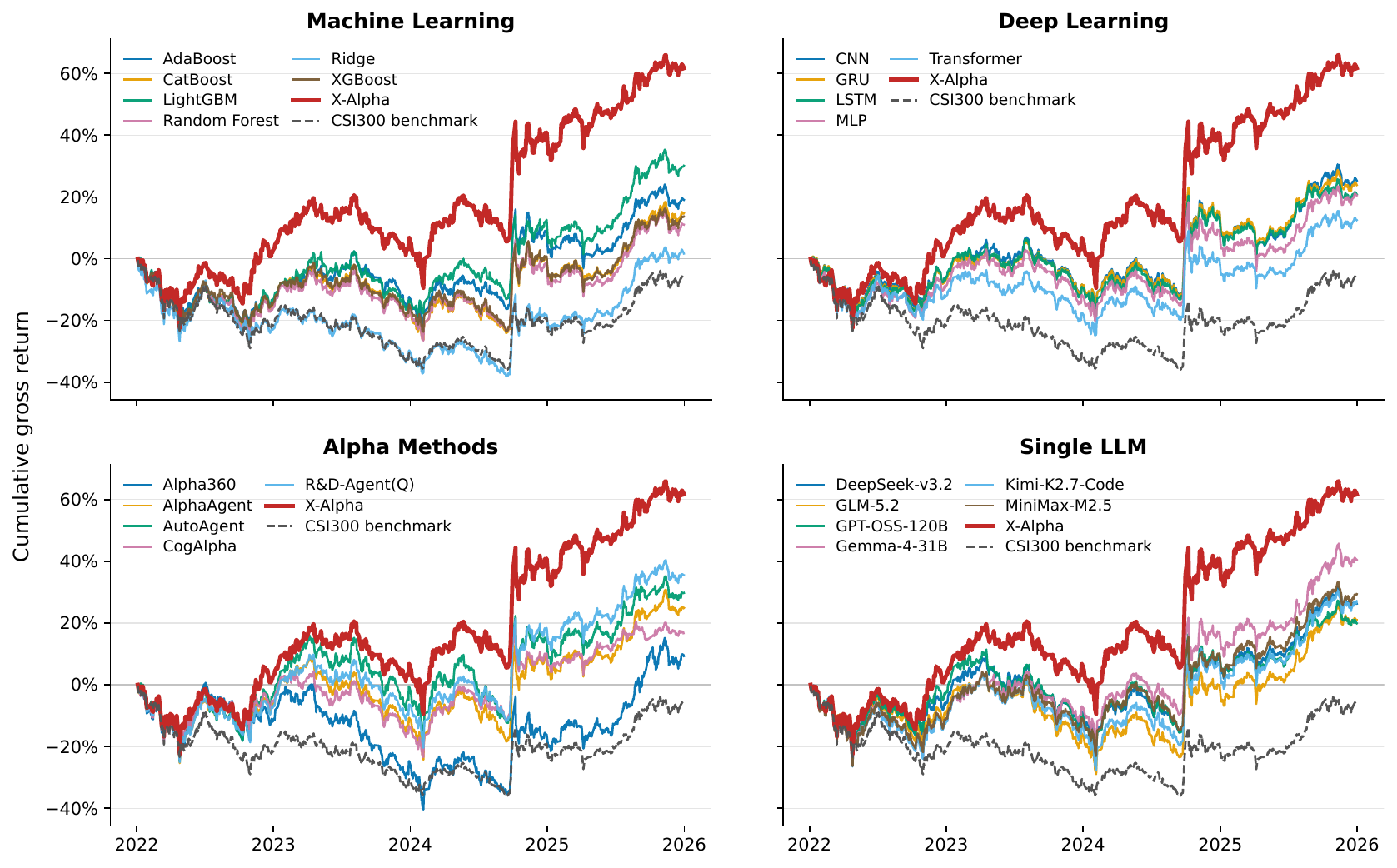}
\caption{CSI300 10-day static-fit cumulative gross returns.
Curves average the separately compounded gross returns of five outcomes per method.}
\label{fig:baseline_static_cumulative_returns}
\end{figure}

\subsection{Static-Fit Comparison}
\label{app:baseline_static_results}

Table~\ref{tab:complete_baseline_static} uses the same methods, seeds, factor groups, and aggregation rule as the rolling comparison.
Model fitting follows the static protocol in Appendix~\ref{app:experimental_protocol}.
Figure~\ref{fig:baseline_static_cumulative_returns} shows the corresponding cumulative gross returns.

\subsection{Factor-Selection Sensitivity}
\label{app:xalpha_selection_sensitivity}

We evaluate \method{} factor combinations with factor-count limits of 10, 20, 30, 40, and 50 and absolute correlation thresholds of 0.6, 0.7, 0.8, and 0.9.
Each configuration uses one library-score-ranked factor group fitted with Ridge under the static protocol.
Table~\ref{tab:xalpha_selection_sensitivity} reports its predictive and portfolio metrics.

\begin{table*}[!htbp]
\centering
\scriptsize
\setlength{\tabcolsep}{4pt}
\renewcommand{\arraystretch}{1.04}
\begin{tabular*}{\textwidth}{@{\extracolsep{\fill}}ccrrrrrrr@{}}
\toprule
\textbf{Max. factors} & \textbf{Corr. threshold} & \textbf{IC} & \textbf{RankIC} & \textbf{ICIR} & \textbf{RankICIR} & \textbf{AR} & \textbf{AER} & \textbf{IR} \\
\midrule
\multirow{4}{*}{10} & 0.6 & \textbf{0.0596} & \textbf{0.0754} & 0.3072 & \textbf{0.3630} & \textbf{0.1328} & 0.1268 & 1.1481 \\
 & 0.7 & 0.0550 & 0.0652 & \textbf{0.3105} & 0.3383 & 0.1297 & \textbf{0.1274} & \textbf{1.2089} \\
 & 0.8 & 0.0481 & 0.0621 & 0.2616 & 0.3150 & 0.0886 & 0.0861 & 0.8201 \\
 & 0.9 & 0.0504 & 0.0692 & 0.2490 & 0.3278 & 0.1065 & 0.1076 & 0.9231 \\
\midrule
\multirow{4}{*}{20} & 0.6 & \textbf{0.0664} & \textbf{0.0829} & \textbf{0.3337} & \textbf{0.3975} & \textbf{0.1969} & \textbf{0.1903} & \textbf{1.7611} \\
 & 0.7 & 0.0546 & 0.0702 & 0.3079 & 0.3770 & 0.1717 & 0.1679 & 1.6058 \\
 & 0.8 & 0.0532 & 0.0689 & 0.2894 & 0.3495 & 0.1323 & 0.1276 & 1.2002 \\
 & 0.9 & 0.0525 & 0.0680 & 0.2844 & 0.3431 & 0.1322 & 0.1288 & 1.1738 \\
\midrule
\multirow{4}{*}{30} & 0.6 & \textbf{0.0636} & 0.0749 & \textbf{0.3417} & \textbf{0.3931} & 0.1621 & 0.1547 & 1.5144 \\
 & 0.7 & 0.0584 & \textbf{0.0759} & 0.3111 & 0.3857 & \textbf{0.1685} & \textbf{0.1643} & \textbf{1.5670} \\
 & 0.8 & 0.0572 & \textbf{0.0759} & 0.3059 & 0.3881 & 0.1628 & 0.1571 & 1.5571 \\
 & 0.9 & 0.0497 & 0.0630 & 0.2677 & 0.3207 & 0.1224 & 0.1163 & 1.0834 \\
\midrule
\multirow{4}{*}{40} & 0.6 & \textbf{0.0608} & 0.0719 & \textbf{0.3723} & \textbf{0.4016} & \textbf{0.1752} & \textbf{0.1667} & \textbf{1.7455} \\
 & 0.7 & 0.0526 & 0.0673 & 0.2779 & 0.3407 & 0.1403 & 0.1333 & 1.2416 \\
 & 0.8 & 0.0581 & \textbf{0.0774} & 0.3036 & 0.3886 & 0.1618 & 0.1545 & 1.5127 \\
 & 0.9 & 0.0489 & 0.0608 & 0.2730 & 0.3255 & 0.1029 & 0.0977 & 0.9232 \\
\midrule
\multirow{4}{*}{50} & 0.6 & 0.0526 & 0.0646 & 0.3244 & 0.3680 & 0.1497 & 0.1411 & 1.4860 \\
 & 0.7 & \textbf{0.0552} & 0.0689 & \textbf{0.3315} & \textbf{0.3717} & \textbf{0.1764} & \textbf{0.1677} & \textbf{1.6745} \\
 & 0.8 & 0.0506 & \textbf{0.0702} & 0.2664 & 0.3585 & 0.1622 & 0.1545 & 1.5878 \\
 & 0.9 & 0.0532 & 0.0687 & 0.2872 & 0.3601 & 0.1479 & 0.1415 & 1.3767 \\
\bottomrule
\end{tabular*}
\caption{\method{} static-fit factor-selection results on CSI300 with a 10-day horizon.
Rows vary the factor-count limit and maximum absolute pairwise correlation.
The best values within each factor-count group are bolded.}
\label{tab:xalpha_selection_sensitivity}
\end{table*}

\section{Mechanism Evaluation Protocols}
\label{app:ablation_protocols}

The seven CSI300 10-day studies cover generation quality, research-record consistency, signal diversity, and candidate selection.
Each study specifies its comparison design and candidate pool.
Selection uses training and validation evidence, with test metrics reserved for evaluation.
Relative gains use unrounded means and are computed as $100(m_{\mathrm{condition}}-m_{\mathrm{reference}})/m_{\mathrm{reference}}$.
The reference is the corresponding control or pre-injection pool.

\paragraph{RMA research-input comparison.}
\label{app:rma_input_comparison}
The RMA condition uses C-layer report-grounded research paths; the control uses \emph{direct report} input.
Both conditions use 2,120 financial reports and follow the same downstream factor-discovery pipeline.
The accepted pools contain 1,417 \emph{direct report} factors and 1,329 RMA-derived factors.
Table~\ref{tab:rma_predictive_quality_full} reports mean test metrics over all accepted factors within each condition, with relative gains summarized in Table~\ref{tab:rma_predictive_quality}.

\begin{table}[!htbp]
\centering
\small
\setlength{\tabcolsep}{3pt}
\begin{tabular*}{\columnwidth}{@{\extracolsep{\fill}}lrrrr@{}}
\toprule
Input & IC & RankIC & ICIR & RankICIR \\
\midrule
\emph{Direct report} & 0.002625 & 0.007485 & 0.009677 & 0.035220 \\
RMA & \textbf{0.003840} & \textbf{0.007964} & \textbf{0.022943} & \textbf{0.045701} \\
Gain & +46.28\% & +6.40\% & +137.09\% & +29.76\% \\
\bottomrule
\end{tabular*}
\caption{Mean held-out metrics over all accepted single factors on CSI300 (10-day).
Gain is relative to \emph{direct report} input.
Better metric values are bolded.}
\label{tab:rma_predictive_quality_full}
\end{table}

The complementary reporting comparison uses correlation limits $\{0.6,0.7,0.8\}$ and maximum subset sizes $\{20,40,60,100\}$.
Each condition's factors are ranked by train--validation library score and selected under the corresponding correlation constraint before Ridge fitting and test evaluation.
Table~\ref{tab:rma_combination_grid} reports both conditions for all 12 configurations.
RMA improves RankIC and RankICIR in every configuration, and AER and IR in nine configurations.
The single-factor results support the effectiveness of RMA in providing more actionable research input and improving the predictive quality of generated factors.
The combination results provide complementary evidence across reporting configurations.

\begin{table*}[!htbp]
\centering
\scriptsize
\setlength{\tabcolsep}{3.2pt}
\renewcommand{\arraystretch}{1.04}
\resizebox{\textwidth}{!}{
\begin{tabular}{@{}cclrrrrrrr@{}}
\toprule
\textbf{Corr.} & \textbf{Max. factors} & \textbf{Input} & \textbf{IC} & \textbf{RankIC} & \textbf{ICIR} & \textbf{RankICIR} & \textbf{AR} & \textbf{AER} & \textbf{IR} \\
\midrule
\multirow{8}{*}{0.6} & \multirow{2}{*}{20} & \emph{Direct report} & \textbf{0.033551} & 0.051606 & \textbf{0.177874} & 0.268481 & 0.049572 & 0.012558 & 0.150578 \\
 &  & RMA & 0.030086 & \textbf{0.053347} & 0.155788 & \textbf{0.275888} & \textbf{0.086369} & \textbf{0.048092} & \textbf{0.569516} \\
 & \multirow{2}{*}{40} & \emph{Direct report} & \textbf{0.022373} & 0.025778 & \textbf{0.152036} & 0.173405 & 0.059989 & 0.024816 & 0.305317 \\
 &  & RMA & 0.020777 & \textbf{0.041034} & 0.118777 & \textbf{0.235949} & \textbf{0.089253} & \textbf{0.053146} & \textbf{0.638341} \\
 & \multirow{2}{*}{60} & \emph{Direct report} & \textbf{0.028656} & 0.033951 & \textbf{0.207075} & 0.242515 & 0.082233 & 0.048511 & 0.548055 \\
 &  & RMA & 0.027513 & \textbf{0.046491} & 0.161036 & \textbf{0.278506} & \textbf{0.114096} & \textbf{0.077034} & \textbf{0.917147} \\
 & \multirow{2}{*}{100} & \emph{Direct report} & 0.031253 & 0.041784 & \textbf{0.213044} & 0.281944 & 0.099416 & 0.063346 & 0.806859 \\
 &  & RMA & \textbf{0.032644} & \textbf{0.046451} & 0.199467 & \textbf{0.296462} & \textbf{0.141460} & \textbf{0.103274} & \textbf{1.266657} \\
\midrule
\multirow{8}{*}{0.7} & \multirow{2}{*}{20} & \emph{Direct report} & 0.022878 & 0.043783 & 0.123366 & 0.233620 & 0.029083 & -0.006121 & -0.076823 \\
 &  & RMA & \textbf{0.030111} & \textbf{0.049686} & \textbf{0.157518} & \textbf{0.252591} & \textbf{0.072972} & \textbf{0.035166} & \textbf{0.420238} \\
 & \multirow{2}{*}{40} & \emph{Direct report} & 0.014056 & 0.021992 & 0.078423 & 0.124186 & 0.016120 & -0.016062 & -0.174827 \\
 &  & RMA & \textbf{0.023079} & \textbf{0.032744} & \textbf{0.143165} & \textbf{0.195073} & \textbf{0.060588} & \textbf{0.025925} & \textbf{0.298020} \\
 & \multirow{2}{*}{60} & \emph{Direct report} & 0.019730 & 0.025168 & 0.131158 & 0.168662 & \textbf{0.049925} & \textbf{0.017258} & \textbf{0.199919} \\
 &  & RMA & \textbf{0.021535} & \textbf{0.028516} & \textbf{0.137680} & \textbf{0.176927} & 0.032009 & -0.000555 & -0.006607 \\
 & \multirow{2}{*}{100} & \emph{Direct report} & 0.028958 & 0.040744 & 0.188367 & 0.270818 & \textbf{0.114189} & \textbf{0.080042} & \textbf{0.941506} \\
 &  & RMA & \textbf{0.031147} & \textbf{0.046074} & \textbf{0.194445} & \textbf{0.284019} & 0.106891 & 0.069994 & 0.844855 \\
\midrule
\multirow{8}{*}{0.8} & \multirow{2}{*}{20} & \emph{Direct report} & 0.009610 & 0.016570 & 0.054774 & 0.090645 & 0.025946 & -0.006976 & -0.082587 \\
 &  & RMA & \textbf{0.014900} & \textbf{0.031652} & \textbf{0.088443} & \textbf{0.180385} & \textbf{0.050245} & \textbf{0.016910} & \textbf{0.183040} \\
 & \multirow{2}{*}{40} & \emph{Direct report} & 0.016663 & 0.020957 & 0.098289 & 0.123664 & 0.033025 & -0.000208 & -0.002353 \\
 &  & RMA & \textbf{0.024551} & \textbf{0.039057} & \textbf{0.148838} & \textbf{0.235010} & \textbf{0.072030} & \textbf{0.036736} & \textbf{0.440234} \\
 & \multirow{2}{*}{60} & \emph{Direct report} & 0.015911 & 0.021741 & 0.094121 & 0.129776 & 0.033373 & 0.000255 & 0.002898 \\
 &  & RMA & \textbf{0.027665} & \textbf{0.036318} & \textbf{0.193519} & \textbf{0.250355} & \textbf{0.055414} & \textbf{0.021593} & \textbf{0.264948} \\
 & \multirow{2}{*}{100} & \emph{Direct report} & 0.021955 & 0.031515 & 0.141907 & 0.205587 & \textbf{0.081367} & \textbf{0.046277} & \textbf{0.569437} \\
 &  & RMA & \textbf{0.030254} & \textbf{0.038246} & \textbf{0.208483} & \textbf{0.264841} & 0.073918 & 0.039533 & 0.528491 \\
\bottomrule
\end{tabular}
}
\caption{RMA research-input comparison across 12 factor-count/correlation configurations on CSI300 (10-day).
Each row reports test metrics for one correlation-constrained Ridge combination.
Better values within each configuration are bolded.}
\label{tab:rma_combination_grid}
\end{table*}

\begin{figure*}[!htbp]
\centering
\includegraphics[width=0.94\textwidth]{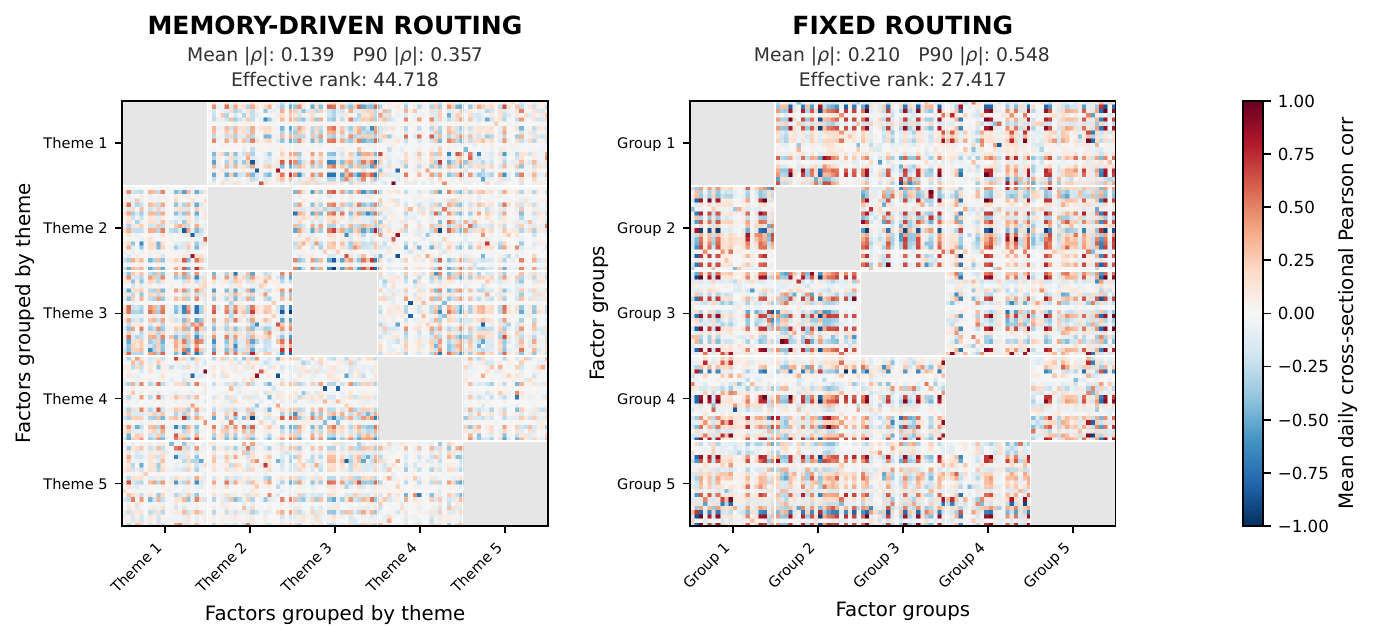}
\caption{Representative cross-cycle signed-correlation matrices for memory-driven and fixed routing.
Same-cycle blocks are masked in gray.
Table~\ref{tab:routing_diversity_full} reports the mean and standard deviation across five independent experiments.}
\label{fig:routing_heatmap_appendix}
\end{figure*}

\paragraph{Cross-cycle routing comparison.}
We compare memory-driven, fixed, and random routing under the same setting.
Memory-driven routing adapts themes using accumulated feedback, and fixed routing retains one theme and archetype bundle across cycles.
Five independent experiments per condition each sample 20 normal-selected factors from the initial generation of each of five cycles (100 factors and 4,000 cross-cycle pairs).
Cross-cycle statistics average daily cross-sectional Pearson signal correlations for pairs from different cycles.
Effective rank is the entropy-based rank of the full 100-factor correlation matrix.
Table~\ref{tab:routing_diversity_full} reports means and sample standard deviations across the five experiments.
Memory-driven routing reduces mean absolute cross-cycle correlation by 35.6\% and increases effective rank by 67.4\% relative to fixed routing.
Random routing yields intermediate diversity between fixed and memory-driven routing.
Figure~\ref{fig:routing_heatmap_appendix} visualizes the cross-cycle correlation contrast.
For predictive quality, we select top-10 and top-20 factors by train--validation RankIC and evaluate their mean held-out RankIC with training-fixed factor directions.
Table~\ref{tab:routing_selected_quality} reports the mean and sample standard deviation across the five independent experiments.
Memory-driven routing improves the top-10 and top-20 test RankIC by 27.9\% and 15.3\%, respectively.
Together, these results support reduced repeated exploration across cycles and greater factor-signal diversity, with broader search yielding a stronger upper tail of selectable high-quality factors.

\begin{table}[t]
\centering
\small
\begin{tabular*}{\textwidth}{@{\extracolsep{\fill}}lrrr@{}}
\toprule
Metric & Fixed & Random & Memory-driven \\
\midrule
Mean $|\mathrm{corr}|$ $\downarrow$ & $0.2106\pm0.0127$ & $0.1980\pm0.0134$ & $\mathbf{0.1357\pm0.0128}$ \\
P90 $|\mathrm{corr}|$ $\downarrow$ & $0.5374\pm0.0192$ & $0.5210\pm0.0216$ & $\mathbf{0.3489\pm0.0289}$ \\
Effective rank $\uparrow$ & $27.01\pm1.66$ & $29.836\pm0.0235$ & $\mathbf{45.22\pm3.44}$ \\
\bottomrule
\end{tabular*}
\caption{Cross-cycle routing comparison across five independent experiments.
Values are mean $\pm$ sample standard deviation; the best mean is bolded.}
\label{tab:routing_diversity_full}
\end{table}

\begin{table}[t]
\centering
\small
\begin{tabular*}{\columnwidth}{@{\extracolsep{\fill}}lrr@{}}
\toprule
Subset & Fixed & Memory-driven \\
\midrule
Top-10 & $0.02059\pm0.00225$ & $\mathbf{0.02634\pm0.00372}$ \\
Top-20 & $0.01599\pm0.00179$ & $\mathbf{0.01843\pm0.00157}$ \\
\bottomrule
\end{tabular*}
\caption{Test RankIC of candidates selected by train--validation RankIC.
Values are mean $\pm$ sample SD across five independent experiments; the better mean is bolded.}
\label{tab:routing_selected_quality}
\end{table}

\paragraph{Discovery feedback controls.}
\label{app:feedback_k_sensitivity}
We conduct two experiments with 100 generation requests per condition, comparing no feedback, real feedback, and random feedback from the same candidate bank.
Real GOOD/BAD feedback is selected by empirical performance and selection scores.
Random feedback samples GOOD/BAD-shaped records from the bank.
Candidates are ranked by train--validation alpha score for equally sized top-$K$ comparisons at $K\in\{5,10,20,50\}$, with test metrics used only for evaluation.
All accepted candidates are also evaluated without top-$K$ truncation.
Table~\ref{tab:discovery_feedback_quality} presents the top-5 test metrics averaged across both experiments.
Table~\ref{tab:discovery_feedback_all_valid} supplements the main result with all-valid metrics for each experiment.
Real feedback leads both controls on both all-valid metrics in each experiment.

Table~\ref{tab:feedback_k_sensitivity} averages both experiments and shows that real feedback obtains the highest test RankIC and RankICIR at every scope.
These results support the effectiveness of historical success and failure feedback in guiding factor generation and improving the predictive quality of subsequent candidates.

\begin{table}[!htbp]
\centering
\small
\begin{tabular}{lrrrrrr}
\toprule
 & \multicolumn{3}{c}{Experiment 1} & \multicolumn{3}{c}{Experiment 2} \\
\cmidrule(lr){2-4}\cmidrule(lr){5-7}
Feedback & Valid & RankIC & RankICIR & Valid & RankIC & RankICIR \\
\midrule
None & 79 & 0.001560 & 0.015202 & 87 & 0.001031 & 0.010239 \\
Random & 62 & 0.001748 & 0.016291 & 65 & 0.002425 & 0.022129 \\
Real & 66 & \textbf{0.010285} & \textbf{0.072807} & 54 & \textbf{0.004837} & \textbf{0.038108} \\
\bottomrule
\end{tabular}
\caption{All-valid discovery-feedback results in two experiments.
Valid denotes the accepted-factor count; RankIC and RankICIR are mean held-out test metrics.}
\label{tab:discovery_feedback_all_valid}
\end{table}

\begin{table}[!htbp]
\centering
\small
\begin{tabular}{llrr}
\toprule
Scope & Feedback & RankIC & RankICIR \\
\midrule
\multirow{3}{*}{Top-5} & None & 0.008456 & 0.063165 \\
 & Random & 0.006767 & 0.057163 \\
 & Real & \textbf{0.042562} & \textbf{0.261827} \\
\midrule
\multirow{3}{*}{Top-10} & None & 0.004225 & 0.033604 \\
 & Random & 0.005683 & 0.049424 \\
 & Real & \textbf{0.028572} & \textbf{0.185949} \\
\midrule
\multirow{3}{*}{Top-20} & None & 0.003149 & 0.026367 \\
 & Random & 0.003988 & 0.035187 \\
 & Real & \textbf{0.017750} & \textbf{0.123952} \\
\midrule
\multirow{3}{*}{Top-50} & None & 0.000753 & 0.009693 \\
 & Random & 0.002684 & 0.024347 \\
 & Real & \textbf{0.009415} & \textbf{0.069520} \\
\midrule
\multirow{3}{*}{All-valid} & None & 0.001295 & 0.012721 \\
 & Random & 0.002087 & 0.019210 \\
 & Real & \textbf{0.007561} & \textbf{0.055458} \\
\bottomrule
\end{tabular}
\caption{Discovery feedback across selection sizes, averaged over two experiments.
Top-$K$ factors are selected using train--validation scores; all-valid uses each condition's accepted pool.
Values are mean single-factor test metrics.}
\label{tab:feedback_k_sensitivity}
\end{table}

\paragraph{Tri-alignment repair audit.}
We randomly sample 167 factor records from 1,500 candidates and compare the same records before and after tri-alignment repair.
Each pair contains the original code and financial explanation and their repaired versions.
Manual review checks whether the explanation agrees with the code.
It also assesses the financial soundness of the code and explanation separately.
A record passes the joint criterion only if all three checks pass.
The number of passing records increases from 21/167 (12.57\%) before repair to 150/167 (89.82\%) after repair.
The comparison supports improved consistency among hypotheses, factor code, and financial explanations, improving the quality of research-memory entries.

\paragraph{Novelty-injection diagnostic.}
We compare parent pools before and after adding screened novelty factors and deduplicating in 10 CSI300 10-day injection events (Table~\ref{tab:novelty_injection_events}).
Mean and P90 absolute pairwise correlations measure pool redundancy, and normalized effective rank divides the signal correlation matrix's effective rank by pool size.
Novelty--parent mean maximum averages each injected factor's highest absolute correlation with the original parents.
P90 is the 90th percentile across all absolute novelty--parent correlations.
The mean row reports event-averaged metrics and relative changes between the before/after averages.
Both parent-pool correlation measures decrease in all 10 events, while normalized effective rank increases in eight.
These results support the effectiveness of novelty injection in reducing factor redundancy during within-cycle evolution.

\begin{table*}[!htbp]
\centering
\small
\setlength{\tabcolsep}{3pt}
\renewcommand{\arraystretch}{1.05}
\textbf{(a) Parent-pool size and normalized diversity}
\par\smallskip
\resizebox{\textwidth}{!}{
\begin{tabular}{@{}rrrrrrrrrrrrr@{}}
\toprule
\multirow{2}{*}{Event} & \multirow{2}{*}{Injected} & \multicolumn{2}{c}{Pool size} & \multicolumn{3}{c}{Mean $|\mathrm{corr}|$ $\downarrow$} & \multicolumn{3}{c}{P90 $|\mathrm{corr}|$ $\downarrow$} & \multicolumn{3}{c}{Normalized effective rank $\uparrow$} \\
\cmidrule(lr){3-4}\cmidrule(lr){5-7}\cmidrule(lr){8-10}\cmidrule(lr){11-13}
 & & Before & After & Before & After & Change & Before & After & Change & Before & After & Change \\
\midrule
1 & 3 & 39 & 42 & 0.3746 & \textbf{0.3346} & -10.67\% & 0.6601 & \textbf{0.6316} & -4.33\% & 0.2647 & \textbf{0.2844} & +7.44\% \\
2 & 4 & 43 & 47 & 0.4831 & \textbf{0.4352} & -9.92\% & 0.7498 & \textbf{0.7373} & -1.67\% & 0.1868 & \textbf{0.2046} & +9.50\% \\
3 & 3 & 45 & 48 & 0.2290 & \textbf{0.2237} & -2.31\% & 0.4450 & \textbf{0.4249} & -4.51\% & 0.3388 & \textbf{0.3419} & +0.91\% \\
4 & 6 & 38 & 44 & 0.2787 & \textbf{0.2777} & -0.35\% & 0.5088 & \textbf{0.4949} & -2.73\% & \textbf{0.3858} & 0.3738 & -3.11\% \\
5 & 3 & 40 & 43 & 0.2669 & \textbf{0.2474} & -7.31\% & 0.6539 & \textbf{0.6154} & -5.89\% & 0.3200 & \textbf{0.3377} & +5.54\% \\
6 & 4 & 46 & 50 & 0.2557 & \textbf{0.2479} & -3.06\% & 0.5720 & \textbf{0.5542} & -3.11\% & 0.3411 & \textbf{0.3476} & +1.90\% \\
7 & 5 & 44 & 49 & 0.2566 & \textbf{0.2456} & -4.26\% & 0.4981 & \textbf{0.4654} & -6.56\% & 0.3736 & \textbf{0.3828} & +2.47\% \\
8 & 4 & 32 & 36 & 0.3371 & \textbf{0.3359} & -0.38\% & 0.6810 & \textbf{0.6663} & -2.15\% & \textbf{0.3428} & 0.3294 & -3.90\% \\
9 & 3 & 40 & 43 & 0.3888 & \textbf{0.3704} & -4.73\% & 0.7593 & \textbf{0.7279} & -4.13\% & 0.2447 & \textbf{0.2537} & +3.71\% \\
10 & 4 & 40 & 44 & 0.3811 & \textbf{0.3415} & -10.40\% & 0.7208 & \textbf{0.6874} & -4.63\% & 0.2340 & \textbf{0.2576} & +10.06\% \\
\midrule
Mean & -- & -- & -- & 0.3252 & \textbf{0.3060} & -5.90\% & 0.6249 & \textbf{0.6005} & -3.90\% & 0.3032 & \textbf{0.3113} & +2.68\% \\
\bottomrule
\end{tabular}
}
\par\medskip
\textbf{(b) Raw effective rank and injected-factor redundancy}
\par\smallskip
\begin{tabular*}{\textwidth}{@{\extracolsep{\fill}}rrrrrr@{}}
\toprule
\multirow{2}{*}{Event} & \multicolumn{3}{c}{Effective rank $\uparrow$} & \multicolumn{2}{c}{Novelty--parent $|\mathrm{corr}|$ $\downarrow$} \\
\cmidrule(lr){2-4}\cmidrule(lr){5-6}
 & Before & After & Change & Mean maximum & P90 \\
\midrule
1 & 10.3242 & \textbf{11.9450} & +15.70\% & 0.3298 & 0.1873 \\
2 & 8.0340 & \textbf{9.6155} & +19.68\% & 0.3471 & 0.4761 \\
3 & 15.2460 & \textbf{16.4106} & +7.64\% & 0.5095 & 0.3906 \\
4 & 14.6590 & \textbf{16.4455} & +12.19\% & 0.5220 & 0.4601 \\
5 & 12.7990 & \textbf{14.5207} & +13.45\% & 0.3310 & 0.2396 \\
6 & 15.6887 & \textbf{17.3775} & +10.76\% & 0.4815 & 0.3983 \\
7 & 16.4369 & \textbf{18.7573} & +14.12\% & 0.4177 & 0.3398 \\
8 & 10.9685 & \textbf{11.8586} & +8.11\% & 0.6330 & 0.5920 \\
9 & 9.7867 & \textbf{10.9112} & +11.49\% & 0.4780 & 0.4547 \\
10 & 9.3612 & \textbf{11.3333} & +21.07\% & 0.3509 & 0.3029 \\
\midrule
Mean & 12.3304 & \textbf{13.9175} & +12.87\% & 0.4401 & 0.3841 \\
\bottomrule
\end{tabular*}
\caption{Complete novelty-injection results for 10 selected events on CSI300 (10-day).
Before and after refer to the parent pool immediately surrounding injection.
Changes are relative percentages; normalized effective rank divides effective rank by pool size.
Better values within each before--after pair are bolded.}
\label{tab:novelty_injection_events}
\end{table*}

\paragraph{Rolling OOS selection study.}
We compare selection with and without rolling OOS hard filtering on a fixed bank of 1,417 evaluated CSI300 10-day candidates.
The filter checks rolling-window robustness.
The elite pool contains candidates passing elite selection, and the normal pool contains the remaining normal-selected candidates.
Screening and ranking use train--validation evidence in this fixed-bank study, and held-out test metrics assess the selected single factors.
Table~\ref{tab:rolling_oos_selection} compares None and Filter only in both pools at selection limits $K\in\{40,100,200\}$.

Hard filtering improves both test metrics in the normal and elite pools at all tested selection limits.
At top-200 in the normal pool, mean held-out RankIC and RankICIR improve by 14.56\% and 17.85\%, respectively, relative to None.
These results support the effectiveness of rolling OOS hard filtering in identifying more reliable candidates and improving selection quality.

\begin{table}[!htbp]
\centering
\small
\setlength{\tabcolsep}{3pt}
\renewcommand{\arraystretch}{1.00}
\begin{tabular*}{\linewidth}{@{\extracolsep{\fill}}rllrrrr@{}}
\toprule
\multirow{2}{*}{$K$} & \multirow{2}{*}{Pool} & \multirow{2}{*}{Policy} & \multicolumn{2}{c}{Test RankIC} & \multicolumn{2}{c}{Test RankICIR} \\
\cmidrule(lr){4-5}\cmidrule(lr){6-7}
 & & & Mean & Gain & Mean & Gain \\
\midrule
\multirow{4}{*}{40} & \multirow{2}{*}{Normal} & None & 0.019831 & -- & 0.124032 & -- \\
 &  & Filter only & \textbf{0.021059} & +6.19\% & \textbf{0.132862} & +7.12\% \\
\cmidrule(lr){2-7}
 & \multirow{2}{*}{Elite} & None & 0.063033 & -- & 0.268282 & -- \\
 &  & Filter only & \textbf{0.064237} & +1.91\% & \textbf{0.272735} & +1.66\% \\
\midrule
\multirow{4}{*}{100} & \multirow{2}{*}{Normal} & None & 0.019177 & -- & 0.111157 & -- \\
 &  & Filter only & \textbf{0.020143} & +5.04\% & \textbf{0.120885} & +8.75\% \\
\cmidrule(lr){2-7}
 & \multirow{2}{*}{Elite} & None & 0.056411 & -- & 0.239765 & -- \\
 &  & Filter only & \textbf{0.056670} & +0.46\% & \textbf{0.241951} & +0.91\% \\
\midrule
\multirow{4}{*}{200} & \multirow{2}{*}{Normal} & None & 0.014482 & -- & 0.078548 & -- \\
 &  & Filter only & \textbf{0.016590} & +14.56\% & \textbf{0.092568} & +17.85\% \\
\cmidrule(lr){2-7}
 & \multirow{2}{*}{Elite} & None & 0.051710 & -- & 0.229050 & -- \\
 &  & Filter only & \textbf{0.055133} & +6.62\% & \textbf{0.237014} & +3.48\% \\
\bottomrule
\end{tabular*}
\caption{Rolling OOS selection results on a fixed CSI300 10-day candidate bank.
Metrics are mean held-out single-factor values.
Gains are relative to None within the same pool and selection limit $K$.
Better means within each pool and selection limit are bolded.}
\label{tab:rolling_oos_selection}
\end{table}

\paragraph{Complexity-discount selection study.}
We compare the four complexity-discount options in \method{}---piecewise, linear, exponential, and sigmoid---with no score discount.
All conditions select 40 factors from the same fixed pool of 480 RMA-derived candidates.
Each discount multiplies the ranking score by a complexity-dependent weight bounded below by 0.70.
This limits the complexity penalty to at most a 30\% reduction in the score.
Mean test RankIC and RankICIR and mean AST-based complexity are computed over the selected factors.
All four options improve held-out predictive quality while reducing mean factor complexity (Table~\ref{tab:complexity_quality}).
These results suggest that appropriate complexity control can improve factor quality.

\begin{table}[!htbp]
\centering
\small
\begin{tabular}{lrrr}
\toprule
Discount & RankIC $\uparrow$ & RankICIR $\uparrow$ & Complexity $\downarrow$ \\
\midrule
None & 0.055600 & 0.247191 & 4.4798 \\
Piecewise & 0.058420 & 0.254035 & 4.1525 \\
Linear & 0.058907 & 0.257369 & \textbf{3.9111} \\
Exponential & 0.058828 & 0.257452 & 4.0417 \\
Sigmoid & \textbf{0.060039} & \textbf{0.260331} & 3.9453 \\
\bottomrule
\end{tabular}
\caption{Complexity discounting on a fixed RMA candidate pool: top-40 selection with multiplier floor 0.70.
Predictive metrics are selected single-factor test means; complexity is the mean AST-based complexity score.}
\label{tab:complexity_quality}
\end{table}

\subsection{Computational Cost Aggregation}
\label{app:cost_protocol}

\begin{table}[!htbp]
\centering
\small
\begin{tabular}{lr}
\toprule
Measurement & Mean \\
\midrule
Initialization & 17.52 min \\
Evolutionary generation & 20.69 min \\
Finalization & 35.95 min \\
Mining loop per cycle & $4.00\pm0.18$ h \\
LLM calls per cycle & 8,351 \\
Tokens per cycle & 23.85 million \\
\bottomrule
\end{tabular}
\caption{Representative computational cost from six completed research cycles.
Stage means use 6 initialization stages, 54 evolutionary generations, and 6 finalization stages.
Calls and tokens include all cycle-attributed requests; wall-clock timing covers the mining loop.}
\label{tab:computational_cost}
\end{table}

Table~\ref{tab:computational_cost} summarizes six completed CSI300 10-day cycles using locally deployed \texttt{gpt-oss-120b}.

\section{Generated-Factor Analysis and Case Studies}
\label{app:case_studies}

\subsection{Within-Theme Factor Diversity}

\begin{figure}[t]
\centering
\includegraphics[width=0.52\textwidth]{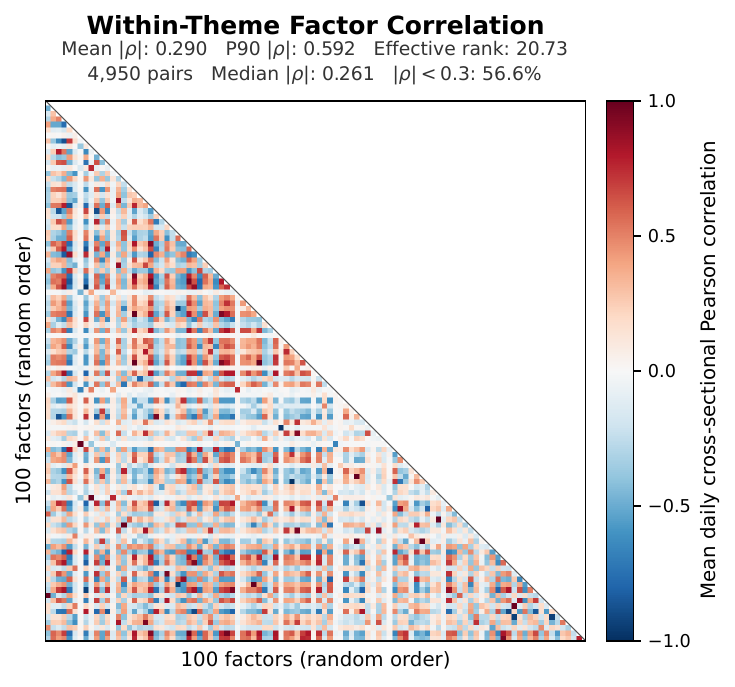}
\vspace{-0.6cm}
\caption{Within-theme factor diversity.
Signed mean daily cross-sectional Pearson correlations among 100 generated factor signals sampled with a fixed seed.
Factors retain their random sample order; only the off-diagonal lower triangle is shown.}
\label{fig:within_theme_factor_corr_heatmap}
\end{figure}

We examine whether hypothesis-to-code discovery collapses into near-duplicate constructions when the research theme is fixed.
Figure~\ref{fig:within_theme_factor_corr_heatmap} reports correlations among 100 factors sampled from one volatility-regime breakout interaction cycle.

Across the 4,950 factor pairs, mean and median absolute correlations are 0.290 and 0.261, respectively; the 90th percentile is 0.592, and 56.6\% of pairs are below 0.3.
The full correlation matrix has an effective rank of 20.73.
These results indicate meaningful variation in signal construction within a shared theme.

\subsection{Representative Factor Code}
\label{app:representative_factor_code}

The representative elite CSI300 factor encodes the hypothesis that reversal following downside overshoot occurs faster in high-volatility regimes.
Its implementation measures downside overshoot relative to a 20-day moving-average anchor and accumulates pressure weighted by absolute returns.
It uses an EWMA span of 5 in high-volatility states and 15 otherwise to represent regime-dependent decay of this pressure.
The factor achieves Train--Val RankIC $=0.0564$ and held-out RankIC $=0.0634$.
Listing~\ref{lst:xalpha_factor_full} provides the code.

\begin{lstlisting}[caption={Implementation excerpt for the regime-dependent overshoot-pressure factor}, label={lst:xalpha_factor_full}, numbers=none]
def regime_overshoot_pressure_decay_20d_ma20_vol20(df):
    """Idea: Downside overshoot pressure decays at a regime-dependent speed,
    capturing faster reversal after capitulation in high-volatility markets.

    Logic: Build a 20-day moving-average anchor and a volatility regime, flag
    downside overshoots with regime-specific thresholds, weight overshoots by
    absolute returns, smooth pressure with fast/slow EWMA decay, and center the
    result by an expanding historical benchmark.

    Sub-mechanism tags: post_capitulation_repair; volatility_regime;
    dynamic_decay; overshoot_weighted.
    """
    close = df["close"]

    # Price anchor and volatility regime.
    ma20 = close.rolling(20, min_periods=10).mean()
    ret = close.pct_change()
    vol20 = ret.rolling(20, min_periods=10).std()
    vol_median = vol20.rolling(60, min_periods=30).median()
    high_vol = (vol20 > vol_median).astype(float)

    # Downside overshoot is stricter in high-volatility states.
    threshold = ma20 * (0.90 * high_vol + 0.95 * (1.0 - high_vol))
    overshoot = (close < threshold).astype(float)

    # Accumulate overshoot pressure and decay it by regime.
    pressure = (overshoot * ret.abs().fillna(0.0)).rolling(20, min_periods=1).sum()
    slow_decay = pressure.ewm(span=15, adjust=False, min_periods=1).mean()
    fast_decay = pressure.ewm(span=5, adjust=False, min_periods=1).mean()
    decayed = high_vol * fast_decay + (1.0 - high_vol) * slow_decay

    return decayed - decayed.expanding().quantile(0.75)
\end{lstlisting}

The factor in Listing~\ref{lst:dynamic_range_turnover_example} combines a dynamic range-burst trigger with turnover excess relative to a median baseline, then smooths the resulting event intensity.
With direction fixed from training data, the factor achieves Train--Val RankIC $=0.0547$ and held-out RankIC $=0.0652$ on CSI300 at the 10-day horizon.

\begin{lstlisting}[caption={Implementation excerpt for the dynamic range-turnover factor}, label={lst:dynamic_range_turnover_example}, numbers=none]
def dynamic_range_turnover_excess_ewm10_factor(df):
    """Idea: Adaptive range-burst detection combined with turnover excess
    relative to a robust median baseline.

    Logic: Measure intraday range against prior close, set a volatility-adaptive
    burst threshold from 30-day log-return volatility, compute log dollar-volume
    excess over a 60-day rolling median, multiply burst intensity by turnover
    excess, and smooth the signal with a 10-day EMA.

    Sub-mechanism tags: range_burst; turnover_excess; volatility_regime;
    ema_smoothing.
    """
    # Range burst threshold adapts to recent close-to-close volatility.
    range_pct = (df["high"] - df["low"]) / df["prev_close"]
    log_ret = np.log(df["close"] / df["prev_close"])
    vol30 = log_ret.rolling(30, min_periods=15).std()
    burst = (range_pct >= 0.05 + 0.5 * vol30).astype(float)

    # Turnover is measured relative to a robust 60-day median baseline.
    turnover_med = df["log_dollar_volume"].rolling(60, min_periods=20).median()
    turnover_excess = (df["log_dollar_volume"] - turnover_med).fillna(0.0)

    # Smooth range bursts confirmed by unusually high turnover.
    signal = burst * turnover_excess
    return signal.ewm(span=10, adjust=False).mean()
\end{lstlisting}

\subsection{Tri-Alignment and Revision}
We illustrate the ex-ante alignment pipeline using a generated factor: \texttt{signflip\_volume\_spike\_lowvol\_bodysign\_30d\_20d}.
The original hypothesis stated that, after a candle-body direction flip, a volume surge in a low-volatility regime predicts continuation in the new direction.
The executable code implemented the same idea by multiplying the body-sign direction, a sign-flip indicator, a low-volatility indicator, and a raw volume ratio:
\begin{equation}
\mathrm{signal}_t
=
\mathrm{sign}(\mathrm{body}_t)
\cdot
\mathbf{1}\{\mathrm{flip}_t\}
\cdot
\frac{\mathrm{volume}_t}{\mathrm{median}_{30}(\mathrm{volume})_t}
\cdot
\mathbf{1}\{\mathrm{TR}_t < 0.8\,\mathrm{median}_{20}(\mathrm{TR})_t\}.
\end{equation}
The first alignment review accepted the text-code correspondence and the explanatory financial rationale, but rejected the stripped executable code as financially unsound.
The uncapped volume ratio can let extreme volume observations dominate the signal, undermining the intended ``volume surge under low volatility'' mechanism.

\begin{center}
\small
\begin{tabular}{lcc}
\toprule
Alignment check & Before revision & After revision \\
\midrule
Explanatory text--code alignment & Accept & Accept \\
Stripped-code financial soundness & Reject & Accept \\
Explanatory-text financial soundness & Accept & Accept \\
\bottomrule
\end{tabular}
\end{center}

The revision preserved the same event structure, but replaced the raw volume multiplier with a capped relative-volume term and updated the hypothesis and logic text accordingly.
The before/after fragments below show only the mechanism-changing lines:
\begin{lstlisting}[language=Python, caption={Before alignment revision: uncapped volume term}, numbers=none]
volume_ratio = df_copy["volume"] / df_copy["volume"].rolling(30, min_periods=10).median()
volume_ratio = volume_ratio.replace([np.inf, -np.inf], np.nan).fillna(1.0)

factor = body_sign * volume_ratio * low_vol_regime
factor = factor.where(sign_flip, other=0.0)
\end{lstlisting}

\begin{lstlisting}[language=Python, caption={After alignment revision: capped volume term}, numbers=none]
volume_ratio = df_copy["volume"] / df_copy["volume"].rolling(30, min_periods=10).median()
volume_ratio = volume_ratio.replace([np.inf, -np.inf], np.nan).fillna(1.0)
volume_ratio = np.minimum(volume_ratio, 3.0)

factor = body_sign * volume_ratio * low_vol_regime
factor = factor.where(sign_flip, other=0.0)
\end{lstlisting}
The hypothesis and logic were updated to describe the capped volume multiplier.
A second alignment review then accepted all three checks.
This case illustrates how tri-alignment identifies a financially unstable implementation and guides a revision that preserves the intended mechanism.

\subsection{Single-Factor Evolution}
The following mutation illustrates a change in a factor's financial mechanism.
The parent factor, \texttt{volstress\_intraday\_ewm5\_vol20}, expressed same-day volume stress: current volume was divided by its 20-day mean, multiplied by same-day intraday return, and smoothed with a 5-day EWMA.
The mutation operator preserved the volume-stress and intraday-momentum mechanism, but changed the behavioral assumption from same-day stress amplification to lagged pressure response.
It also replaced the 20-day mean volume baseline with a 20-day median baseline to make the stress proxy less sensitive to extreme volume observations.

\begin{lstlisting}[language=Python, caption={Parent factor before mutation: same-day volume stress}]
volume_stress = df["volume"] / df["volume"].rolling(20, min_periods=20).mean()
same_day_pressure = volume_stress * df["intra_1"]
factor = same_day_pressure.ewm(span=5, adjust=False).mean()
\end{lstlisting}

\begin{lstlisting}[language=Python, caption={Mutated child factor: lagged median-based volume stress}]
volume_stress = df["volume"] / df["volume"].rolling(20, min_periods=20).median()
lagged_pressure = volume_stress.shift(1) * df["intra_1"]
factor = lagged_pressure.ewm(span=5, adjust=False).mean()
\end{lstlisting}

The resulting child factor has the hypothesis: prior-day median-based volume stress modulates same-day intraday returns, capturing lagged directional pressure.
The child passed all three tri-alignment checks.

\section{Primitive Feature Definitions}
\label{app:primitive_features}

Table~\ref{tab:primitive-feature-definitions} defines the five daily OHLCV fields and 20 derived features available to factor code.
Derived features are computed separately for each stock in chronological order using current and historical observations.
Let $O_t$, $H_t$, $L_t$, $C_t$, and $V_t$ denote the preprocessed OHLCV values, with $R_t=H_t-L_t$.
Volume scaling and preprocessing are specified in Appendix~\ref{app:experimental_protocol}.

\begingroup
\small
\setlength{\tabcolsep}{5pt}
\renewcommand{\arraystretch}{1.12}
\begin{longtable}{>{\raggedright\arraybackslash}p{0.43\linewidth}%
                  >{\raggedright\arraybackslash}p{0.51\linewidth}}
\toprule
\textbf{Feature} & \textbf{Definition} \\
\midrule
\endfirsthead
\toprule
\textbf{Feature} & \textbf{Definition} \\
\midrule
\endhead
\midrule
\multicolumn{2}{r}{\footnotesize Continued on next page} \\
\endfoot
\bottomrule
\caption{Primitive feature definitions.}
\label{tab:primitive-feature-definitions}\\
\endlastfoot
\multicolumn{2}{l}{\textit{Daily OHLCV fields}} \\
\addlinespace[2pt]
\path{open} & Adjusted opening price, $O_t$. \\
\path{high} & Adjusted daily high, $H_t$. \\
\path{low} & Adjusted daily low, $L_t$. \\
\path{close} & Adjusted closing price, $C_t$. \\
\path{volume} & Preprocessed daily trading volume, $V_t$. \\
\addlinespace[4pt]
\multicolumn{2}{l}{\textit{Derived OHLCV primitive features}} \\
\addlinespace[2pt]
\path{prev_close} & Previous trading-session close, $C_{t-1}$. \\
\path{log_close} & Log closing price, $\log C_t$. \\
\path{ret_1} & Daily log return, $\log C_t-\log C_{t-1}$. \\
\path{gap_1} & Overnight log return, $\log(O_t/C_{t-1})$. \\
\path{intra_1} & Intraday log return, $\log(C_t/O_t)$. \\
\path{bar_range} & Daily high--low range, $R_t=H_t-L_t$. \\
\path{true_range} & True range, $\mathrm{TR}_t=\max\{R_t,\allowbreak |H_t-C_{t-1}|,\allowbreak |L_t-C_{t-1}|\}$. \\
\path{dollar_volume} & Turnover proxy, $C_tV_t$, in preprocessed price--volume units. \\
\path{range_pct_1d_prev_close} & Relative daily range, $R_t/C_{t-1}$. \\
\path{true_range_pct_1d_prev_close} & Relative true range, $\mathrm{TR}_t/C_{t-1}$. \\
\path{close_location_range_1d_ohlc} & Close location, $(2C_t-H_t-L_t)/R_t$, ranging from $-1$ to $1$ for valid OHLC bars. \\
\path{body_to_range_1d_ohlc} & Signed candle-body fraction, $(C_t-O_t)/R_t$. \\
\path{upper_shadow_to_range_1d_ohlc} & Upper-shadow fraction, $S_t^U/R_t$, where $S_t^U=H_t-\max\{O_t,C_t\}$. \\
\path{lower_shadow_to_range_1d_ohlc} & Lower-shadow fraction, $S_t^L/R_t$, where $S_t^L=\min\{O_t,C_t\}-L_t$. \\
\path{shadow_imbalance_1d_ohlc} & Lower-minus-upper shadow fraction, $(S_t^L-S_t^U)/R_t$. \\
\path{log_volume} & Log-scaled volume, $\log(1+\max\{V_t,0\})$. \\
\path{log_dollar_volume} & Log-scaled turnover proxy, $\log(1+\max\{C_tV_t,0\})$. \\
\path{hl_mid_price} & High--low midpoint, $(H_t+L_t)/2$. \\
\path{oc_mid_price} & Open--close midpoint, $(O_t+C_t)/2$. \\
\path{close_to_open_ratio} & Intraday simple return, $C_t/O_t-1$. \\
\end{longtable}
\endgroup

Zero denominators are treated as missing, and lagged features are undefined when the required prior observation is unavailable.
Log-price and log-return calculations clip input prices below at $\epsilon=10^{-12}$.

\section{Selected Agent Prompt Excerpts}
\label{app:prompt_excerpts}

This appendix presents key prompt excerpts for research routing, factor discovery, and feedback consolidation.
Uppercase placeholders denote runtime inputs; ellipses mark omitted content.
Report-abstraction prompts and responses are presented in Appendix~\ref{app:rma_details}.
We will publicly release the complete prompt templates upon acceptance.

\subsection{Research Planning and Routing}

\paragraph{Research direction selection.}
The routing prompt uses accumulated cycle feedback to identify an information gap and define the next theme with its relevant mechanism families.

\begin{xpromptlisting}{Research Direction Selection}
Act as an evidence-based research router.

{ROUTING_TASK_BLOCK}
{RECENT_AGENT_CYCLE_SUMMARIES_BLOCK}

Core task:
1. Follow the routing task to identify one concrete mechanism-level information gap.
2. Create a concise cycle theme around that gap.
3. Select one primary B-layer for the main mechanism line.
4. Select {SUPPORTING_B_LAYER_MIN_COUNT}-{SUPPORTING_B_LAYER_MAX_COUNT} supporting B-layers that serve the same gap.

Information-gap rules:
- `feedback_basis.selected_information_gap` must justify `cycle_theme_name`, `cycle_theme_summary`, and every selected B-layer.
- Do not locally polish recent positive mechanisms.
- Do not revisit failed mechanism families unless the theme explicitly avoids their failed assumptions.

B-layer rules:
...
- Choose only documented B-layer ids from: {B_ID_OPTIONS}
- Each supporting B-layer must serve the selected information gap with one distinct function:
  - complement: adds another information dimension
  - contrast: provides an opposing or discriminating view
  - boundary_condition: supplies state conditions or validity boundaries
  - failure_avoidance: avoids failed assumptions from BAD memory
  - extension: extends the primary mechanism into adjacent but different space
...

Required JSON schema:
{
{ROUTING_THEME_SCHEMA_FIELDS}
  "feedback_basis": {
    "well_explored_mechanisms": "mechanism families already explored",
    "failure_patterns_to_avoid": "failed assumptions or dead-end patterns",
    "selected_information_gap": "specific underexplored mechanism-level gap"
  },
  "b_layers": [
    {
      "layer": "<one b_id from Available b_id options>",
      "role": "primary",
      "supporting_function": "primary_line",
      "b_layer_selection_logic": "why this B-layer is the main line for the selected gap"
    },
    ...
  ]
}

Documented B-Layer Taxonomy:
{B_LAYER_TAXONOMY}
\end{xpromptlisting}

\paragraph{Theme-conditioned hypothesis planning.}
The following excerpts show separate calls for research-path planning and hypothesis generation, with retrieval described in Appendix~\ref{app:memory_routing}.

\begin{xpromptlisting}{Research-Path Planning and Hypothesis Generation}
Research-path planning call (excerpt):
Generate C-layer research directions after B-layer allocation.

For every provided candidate C-layer Research Archetype record, return exactly 2 target_research_paths.

Each target_research_path must:
- align with the cycle theme
- align with the archetype's B-layer allocation
- reflect the B-layer `supporting_function`
- align with the archetype's own core mechanism
- describe a mechanism-level information gap or market behavior direction
- be expressible using daily OHLCV information
- be a one-sentence mechanism direction, not a factor blueprint
...

Cycle theme:
- cycle_theme_name: {CYCLE_THEME_NAME}
- cycle_theme_summary: {CYCLE_THEME_SUMMARY}
Selected B-layer plans:
{B_LAYER_PLANS}
Candidate C-layer Research Archetype records:
{CANDIDATE_AGENTS}

Hypothesis generation call (specific-mode excerpt):
Generate `factor_hypotheses` for the provided agent specification.

Rules:
...
- Use only daily OHLCV-observable information.
- Use sub_mechanism_tags to keep hypotheses mutually different; avoid near-duplicate phrasings.
...
- Each hypothesis_idea must be specific, testable, and implementation-leaning.
...

Return ONLY valid JSON:
{
  "factor_hypotheses": [
    {
      "hypothesis_idea": "...",
      "sub_mechanism_tags": ["...", "..."]
    }
  ]
}

Input:
<agent_spec>
Agent Name:
{AGENT_NAME}
Core Mechanism:
{CORE_MECHANISM}
Research Paths:
{RESEARCH_PATHS}
</agent_spec>
\end{xpromptlisting}

\subsection{Hypothesis-to-Code Factor Discovery}

\paragraph{Factor generation.}
\begin{xpromptlisting}{Hypothesis-to-Code Generation}
### Factor Design Guidance: {AGENT_NAME}
- Stay within the assigned C-layer mechanism: {MECHANISM}.
[HYPOTHESES]
...

Generation Task:
- Generate exactly one new factor function in this response.
- The hypotheses above are optional inspiration, not a hard specification.
...
- If the final implementation departs from the input hypotheses, that is acceptable. In that case, make sure `hypothesis_idea` and `logic` describe the final implemented mechanism rather than restating the input wording verbatim.
...

Complexity control (excerpt):
Keep the factor simple, inspectable, and mechanism-driven.
...
- Use one coherent mechanism only.
...
- Every component must have a clear mechanism role.
...

Output Format:
Return exactly one valid JSON object and nothing else.
Required keys:
- `hypothesis_idea`: one concise sentence describing the implemented mechanism.
- `sub_mechanism_tags`: 1-4 concise tags consistent with the final code.
- `logic`: concise explanation of how the code realizes the idea.
- `code`: exactly one Python function.
...
\end{xpromptlisting}

\paragraph{Tri-alignment review.}
\begin{xpromptlisting}{Tri-Alignment Review}
Evaluate whether the factor's explanatory text and executable implementation are aligned and financially sound.

Definitions:
- Explanatory text includes the function docstring and inline comments.
- Stripped code means executable code after removing docstring and comments.

Judge three things:
1. explanatory_text_code_alignment:
   Accept only if the explanatory text accurately describes what the stripped code actually computes.
2. stripped_code_financial_soundness:
   Accept only if the executable code expresses a coherent, causal, daily-OHLCV alpha mechanism and is free of obvious leakage or pseudo-signal construction.
3. explanatory_text_financial_soundness:
   Accept only if the explanatory text describes a financially coherent mechanism without overclaiming or adding mechanisms absent from the code.

### Stripped Code
{STRIPPED_CODE}

### Explanatory Text
{DOCSTRING}
{COMMENTS}
...

Required JSON schema:
{
  "explanatory_text_code_alignment": "Accept/Reject",
  "stripped_code_financial_soundness": "Accept/Reject",
  "explanatory_text_financial_soundness": "Accept/Reject"
}
\end{xpromptlisting}

\paragraph{Novelty injection.}
The fresh-seed brief adds recent GOOD/BAD feedback to the factor-generation context.

\begin{xpromptlisting}{Feedback-Guided Fresh-Seed Generation}
### Fresh Seed Brief
Use this feedback as mechanism guidance, not as formula templates.

#### Reusable Mechanism Principles
{GOOD_FEEDBACK_ROWS}

#### Failure Patterns To Avoid
{BAD_FEEDBACK_ROWS}

### Fresh Seed Design Task
Generate a fresh seed factor within the assigned C-layer mechanism.
Constraints:
- Main novelty must be mechanism-level, not formula-level.
- Prefer a new event definition, conditioning context, response regime, or horizon proxy.
- Avoid obvious local variants of known failed patterns.
- Reuse OHLCV columns only if they serve a different causal role.
- Use only columns listed in the schema.
- Do not invent unavailable external data or fake OHLCV proxies.
...
\end{xpromptlisting}

\subsection{Feedback and Memory Update}

The following excerpts show separate GOOD/BAD feedback calls and cycle-level consolidation.

\begin{xpromptlisting}{Mechanism Feedback and Cycle-Level Memory}
GOOD feedback call (excerpt):
Analyze GOOD feedback records as positive mechanism memory.

For each GOOD feedback record, extract:
- mechanism_family: concise snake_case mechanism-family label
- positive_evidence: mechanism hypothesis supported by the metrics, stated without formulas, exact windows, thresholds, or parameterized transforms
- reusable_principle: causal mechanism lesson that can guide a different factor design without copying implementation details
- avoid_copying: abstract surface pattern not to copy; describe the wrapper or construction class without exact parameters, spans, thresholds, operators, or function names
...

BAD feedback call (excerpt):
Analyze BAD feedback records as negative mechanism memory.

For each BAD feedback record, extract:
- mechanism_family: concise snake_case mechanism-family label
- failure_type: one of weak_standalone_proxy, noisy_binary_event, horizon_mismatch, over_complex_stack, redundant_duplicate, unstable_regime, bad_transform
- failed_assumption: mechanism assumption that likely failed, stated without formulas, exact windows, thresholds, or parameterized transforms
- avoidance_rule: mechanism class or design assumption future generation should avoid, not a parameter-tuning instruction
- repair_condition: mechanism-level evidence or structure required before retrying; otherwise exactly "not worth retrying without a clear repair"
...

Factor feedback records:
{FEEDBACK_RECORDS}

Cycle-summary call (excerpt):
Summarize one completed evolve cycle as compact cycle-level routing memory.

You are given:
- cycle_theme_name
- cycle_theme_summary
- selected_agent_names
- Good feedback positive mechanism memory
- Bad feedback negative mechanism memory
...

Goal:
Produce cycle-level memory for future routing.
Do not explain individual factors.
Do not replace record-level Good/Bad memory.
Use Good memory only to infer successful cycle-level patterns.
Use Bad memory only to infer failed assumptions and dead ends.
...

Required JSON schema:
{
  "cycle_theme_memory": "what mechanism-level information gap this cycle explored",
  "cycle_agent_coverage_memory": "what the selected C-layer Research Archetype records were structurally positioned to cover",
  "cycle_positive_memory": "cycle-level successful mechanism principles inferred from Good memory; do not list factors",
  "cycle_negative_memory": "cycle-level failed assumptions or dead ends inferred from Bad memory; do not list factors"
}

Cycle memory input:
{FACTOR_INFO_PAYLOAD}
\end{xpromptlisting}

%%%%%%%%%%%%%%%%%%%%%%%%%%%%%%%%%%%%%%%%%%%%%%%%%%%%%%%%%%%%

\end{document}